\pdfoutput=1
\documentclass[11pt]{article}

\usepackage[]{EMNLP2022}

\usepackage{times}
\usepackage{latexsym}
\usepackage{hyperref}       % hyperlinks
\usepackage{caption}
\usepackage{subcaption}
\usepackage{amsmath}
\usepackage{amsfonts}
\usepackage{booktabs}
\usepackage{xcolor,colortbl}

\usepackage{tikz}

\def\Snospace~{\S{}}

\usepackage[T1]{fontenc}
\usepackage[utf8]{inputenc}

\usepackage{microtype}

\usepackage[textwidth=0.7in]{todonotes}
\newcommand{\squishlist}{
 \begin{list}{$\circ$}
  { \setlength{\itemsep}{0pt}
     \setlength{\parsep}{3pt}
     \setlength{\topsep}{3pt}
     \setlength{\partopsep}{0pt}
     \setlength{\leftmargin}{1.5em}
     \setlength{\labelwidth}{1em}
     \setlength{\labelsep}{0.5em} } }

\newcommand{\squishend}{
  \end{list}  }

\newcommand{\fact}[1]{\textbf{{#1}}}

\newcommand{\squishenum}{
 \begin{enumerate}{}
  { \setlength{\itemsep}{0pt}
     \setlength{\parsep}{3pt}
     \setlength{\topsep}{3pt}
     \setlength{\partopsep}{0pt}
     \setlength{\leftmargin}{1.5em}
     \setlength{\labelwidth}{1em}
     \setlength{\labelsep}{0.5em} } }

\newcommand{\squishenumend}{
  \end{enumerate}  }

\title{Do Text-to-Text Multi-Task Learners Suffer from Task Conflict?}

\author{David Mueller\textsuperscript{1,2} \hspace{1em} Nicholas Andrews\textsuperscript{1,2} \hspace{1em} Mark Dredze\textsuperscript{1} \\
  \textsuperscript{1}Department of Computer Science, Johns Hopkins University \\
  \textsuperscript{2}Human Language Technology Center of Excellence, Johns Hopkins University \\
  \texttt{\{dam,noa\}@jhu.edu}
  \hspace{1em}
\texttt{mdredze@cs.jhu.edu}
  }

\begin{document}
\maketitle

\begin{abstract}
    Traditional multi-task learning architectures train a single model across multiple tasks
    through a shared encoder followed by task-specific decoders.
	Learning these models
    often requires specialized training algorithms that address task-conflict
	in the shared parameter updates, which otherwise can lead to \emph{negative transfer}.
    A new type of multi-task learning within NLP homogenizes
    multi-task architectures as a shared encoder and language model decoder, which does
    surprisingly well across a range of diverse tasks \cite{T5-raffel2020exploring}.
	Does this new architecture suffer from task-conflicts that require specialized training algorithms?
	We study how certain factors in the shift towards text-to-text models affects
	multi-task conflict and negative transfer,
	finding that both directional conflict and transfer are surprisingly constant across
	architectures.
\end{abstract}

\section{Introduction}

Multi-task learning (MTL) aims to obtain superior results by training a single
model over multiple related tasks~\cite{caruana1997multitask}.
Despite MTL's promises, training neural networks with
a multi-task objective often yields {\em worse} results on a given task than task-specific models, 
resulting in \emph{negative transfer}. 
Why do some MTL settings work while others suffer from negative transfer?
Many attribute this behavior to
optimization challenges in the MTL loss
surface~\citep{mtsurvey-ruder2017overview}, that arise from
\emph{multi-task conflict}: significant differences across the gradients
of different tasks which may trap SGD in poor optima~\citep{pcgrad-yu2020gradient}.
Prior research has focused on methods to
mitigate conflict between tasks during
training to ensure consistent MTL improvements~\citep[][\emph{inter alia}]{gradnorm-pmlr-v80-chen18a,mgda-NIPS2018_7334,pcgrad-yu2020gradient,gradvaccine-wang2020gradient}.

\begin{figure*}[t!]
    \centering
    \includegraphics[scale=0.113,trim=0 400px 0 0]{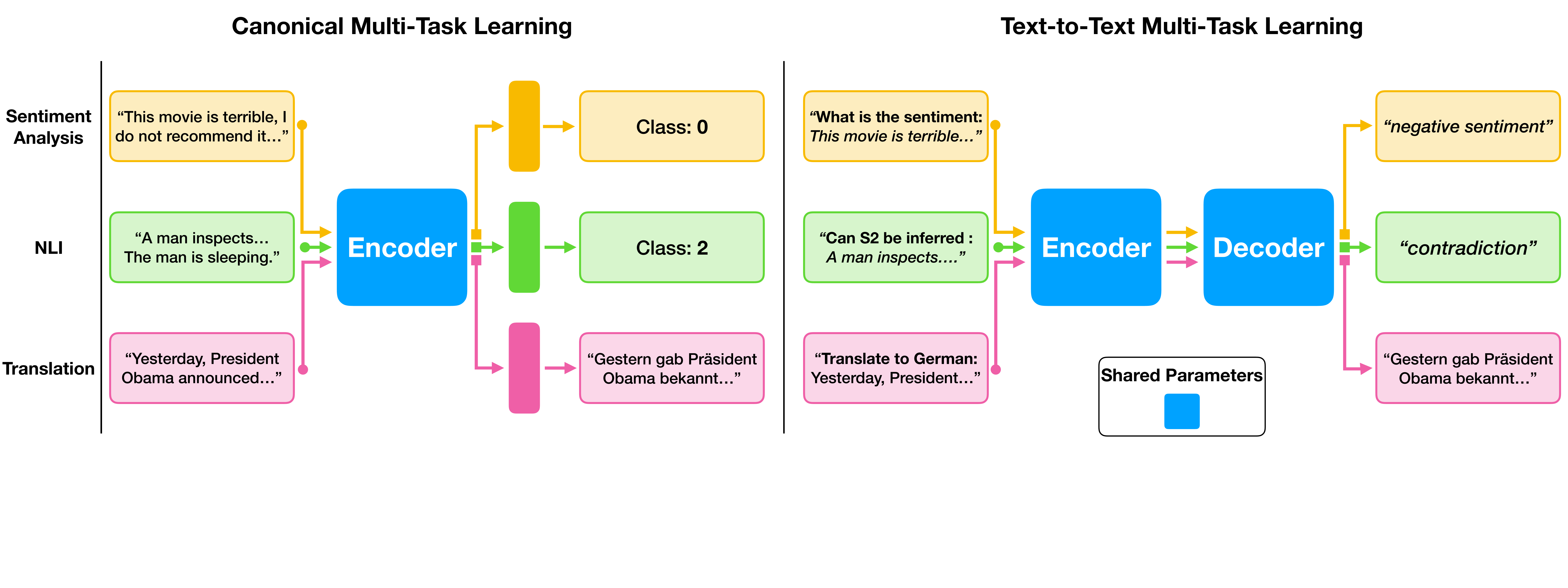}
	\caption{\label{fig:my_label} \emph{Canonical} (left) vs. \emph{Text-to-Text} (right)
		Multi-Task (T5) Architectures.
		In this work we are interested in how \emph{negative transfer} and \emph{multi-task conflict}
		are affected by moving from Canonical MTL to Text-to-Text MTL.}
\end{figure*}

Research on conflict in MTL has largely focused on the {\bf canonical} MTL
setting: all tasks share an \emph{encoder} which
projects inputs into a shared representation space, and then task-predictions
are made using a {\it task-specific head}.
Recently, it has become common in NLP to leverage a
unified {\bf text-to-text} MTL framework: all tasks are
framed as sequence generation problems using a single, fully-shared
language model to make predictions for each
task~\citep[\autoref{fig:my_label};][]{McCann2018decaNLP,T5-raffel2020exploring}.
This framing imposes fewer constraints than the canonical setting, as task-specific output
spaces are relaxed to all natural language sequences, and tasks are inferred by natural language {\it prompts} rather than
architecture.
Despite this flexibility, text-to-text models have achieved state-of-the-art performance on several multi-task benchmarks~\citep{T5-raffel2020exploring,sanh2021multitask}.

In this work, we are interested in the following question: how does reframing MTL with a single text-to-text model affect negative (or positive) transfer, specifically with regard to multi-task conflict in the training objective?
We explore this through the following contributions:
\squishlist
\item We identify three main factors, or model properties, that change between canonical \& text-to-text MTL models which may individually impact multi-task conflict and transfer.
\item We empirically investigate the effect of each of these factors on negative
	transfer and multi-task conflict over two standard MTL benchmarks.
We find that architectural factors affect multi-task conflict but have little effect on transfer.
\item Finally, we show that while task prompts are necessary in multi-task
learning to specify tasks, improving the semantics of task descriptions with
natural language can hurt negative transfer, even if it helps zero-shot capabilities.
\squishend
Overall, our findings suggest that, despite having a fully-shared parameterization between tasks, text-to-text models are surprisingly \textbf{not} inherently superior multi-task learners
than canonical multi-task architectures, and therefore may similarly benefit from research in
conflict mitigation techniques.

\section{Background \& Related Work}
\label{sec:background}

\subsection{Canonical Multi-Task Learning}

Suppose we have $K$ tasks, each with a dataset
$\mathcal{D}_k$ over $\mathcal{X}\times \mathcal{Y}_k$,
where $\mathcal{X}$ is all sequences of tokens from a fixed vocabulary
and $\mathcal{Y}_k$ is a task-specific output space.
Our goal is to learn task-specific models $f_k:\mathcal{X}\rightarrow\mathcal{Y}_k$
for each task while sharing information across all tasks, without any prior 
knowledge about how tasks are related.
Canonically, this is done by training a \emph{shared encoder}
$g_\theta: \mathcal{X} \rightarrow \mathbb{R}^d$ along with a task-specific
head $h_{\phi_k}: \mathbb{R}^d \rightarrow \mathcal{Y}_k$, to project the
input into a shared representation space before predicting into $\mathcal{Y}_k$.\footnote{In this work we are focused on the \emph{hard-sharing} regime
of multi-task learning.
Specifically, we do not consider the \emph{soft-sharing}
regime~\citep{mtsurvey-ruder2017overview}, which shares few to no
parameters between tasks and instead transfers information across
tasks in other ways.}
Parameters $\theta$ are shared across all
tasks, parameters $\phi_k$ are specific to task $k$,
and we specify $f_k(x) \equiv (h_{\phi_k}\circ g_\theta) (x)$.

To train a canonical multi-task model, we first define 
task-specific loss functions $\ell_k(\hat{y}, y)$; for instance, $\ell_k$ may
be mean-squared error for a regression task or cross-entropy for a
classification task.
Let $\mathcal{L}_k(\Theta)$ represent the average task loss 
$\ell_k(f(x;\Theta), y)$ over $(x, y)$ pairs from a mini-batch of $\mathcal{D}_k$,
where $\Theta = \{\theta, \phi_1, \ldots \phi_k\}$ for convenience.
At timestep $t$, we update our parameters with SGD using the gradient update
$\nabla_{\phi_k} \mathcal{L}_k(\Theta^t)$ for all parameters $\phi_k$, and 
$\nabla_\theta^{MT} = \sum_{k=1}^K \nabla_\theta
\mathcal{L}_k(\Theta^t)$
for parameters $\theta$.

Canonical architectures have long been the dominant approach to deep multi-task
learning in deep neural networks~\citep{caruana1997multitask,mtsurvey-ruder2017overview}.
Classically, multi-task conflict occurs in the multi-task gradient update
$\nabla_\theta^{MT}$, and negative transfer arises from difficulties in
learning a proper shared encoder model $g_\theta$.

% \footnote{The parameter update for $\theta$ is occasionally
% writted with a $\frac{1}{K}$ normalizaiton factor over it. We treat this
% normalization factor as a hyper-parameter which we exclude.}

% $\mathcal{L}_k(\Theta)$.
% To do this we define the \emph{multi-task objective} to be the average of each
% task's loss
% 	$\mathcal{L}(\Theta) = \frac{1}{K} \sum_{k\in\mathcal{K}} \mathcal{L}_k (\Theta)$.
% Under this objective, our \emph{shared parameters} $\theta$ have a gradient of
% \begin{equation}
% 	\nabla_\theta \mathcal{L}(\Theta) = \frac{1}{K} \sum_{k \in \mathcal{K}} \nabla_\theta \mathcal{L}_k (\Theta)
% \end{equation}
% which is to say that our shared parameter gradient is the average of each task's
% gradient.

\subsection{Text-to-Text Multi-Task Learning}

% Recently, NLP has seen a significant push towards text-to-text
% multi-task learners: \citet{McCann2018decaNLP} frame multi-task learning as
% question-answering, describing each task as a question and training a single
% Q\&A model to answer questions from all tasks.
% More recently, T5~\citep{T5-raffel2020exploring} and T0~\citep{sanh2021multitask}
% treat multi-task learning as a language modeling problem, prepending or appending
% task prompts to inputs and jointly learning all tasks by fine-tuning a pre-trained
% language model.

In the text-to-text (T2T) MTL setting,  we pre-suppose a function
$r_k:\mathcal{Y}_k \rightarrow \mathcal{X}$
which maps the task-specific output space into natural
language.\footnote{For example, if task $k$ is abstractive summarization, where 
	$\mathcal{Y}_k = \mathcal{X}$, then $r_k$ is simply the identity function.
If task $k$ is a classification task, such as NLI, $r_k$ represents
a mapping of each label into a natural language description.}
Given such an $r_k$ for all tasks, our goal is to learn functions
$f_k:\mathcal{X}\rightarrow\mathcal{X}$ for each task $k$.
Because the output space for each task is the same, we can
make predictions with a \emph{shared decoder}
$j_{\theta_d}: \mathbb{R}^d\rightarrow\mathcal{X}$, which yields a fully
shared encoder-decoder model
$f = j_{\theta_d} \circ g_{\theta_e}: \mathcal{X} \rightarrow \mathcal{X}$,
where $\theta = \{\theta_e, \theta_d\}$ are shared across all tasks and
$g_{\theta_e}$ is defined as above.

Given $f$, how do we specify $f_k$?
A common approach is to prepend a fixed natural language description of the task
$p_k \in \mathcal{X}$ to the input during training and testing, e.g.
$f_k(x) \equiv f(p_k; x)$, where $;$ denotes concatenation.
The hope of this approach is that $p_k$ moves the input into a sufficiently
distinct distribution  that $f$ learns to reliably predict task $k$ for any
input that starts with $p_k$.\footnote{
If the training data from each task comes from sufficiently distinct
distributions, the model may implicitly learn to perform each task based on
the distribution of the input sentence.
However, this may still be undesirable as 
we cannot reliably perform tasks on out-of-distribution inputs.}
This approach has shown surprising success recently:
\citet{McCann2018decaNLP} frame $p_k$ as a task-specific question, and
successfully train a joint Q\&A model to answer every task's questions;
\citet{khashabi-etal-2020-unifiedqa} train a single text-to-text
system on 19 distinct Q\&A tasks, finding that a unified input-output format
is competitive with task-specific formats and systems;
most recently, T5~\citep{T5-raffel2020exploring} and T0~\citep{sanh2021multitask}
treat \textbf{MTL as a language modeling problem}, pre-pending $p_k$
to task inputs and jointly learning all tasks by fine-tuning a pre-trained
encoder-decoder language model.

\subsection{Negative Transfer \& Multi-Task Conflict}
\label{sec:mtl-conflict}

The differentiating factor between multi-task and single-task learning is how
the shared parameters are trained:
$\nabla_\theta^{MT}$ pulls $\theta$ towards a region of the parameter space that
might not be explored by a single-task objective.
Hypothetically, this allows information to transfer across tasks yielding a
stronger model~\citep{caruana1997multitask};
in practice, multi-task objectives often yield {\it negative transfer},
under-performing single-task systems.
Recent work has attributed negative transfer to {\it conflict} between
the directions and magnitudes of task gradients.
When differences between $\nabla_\theta \mathcal{L}_k(\cdot)$ across
different tasks are too large, joint optimization may become difficult.
As a result, prior work has focused on mitigating
conflict by re-weighting task losses
\citep{Kendall_2018_CVPR,gradnorm-pmlr-v80-chen18a,mgda-NIPS2018_7334},
by aligning gradient
directions~\citep{pcgrad-yu2020gradient,gradvaccine-wang2020gradient,NEURIPS2020_16002f7a},
or both~\citep{javaloy2022rotograd}.

Despite progress on mitigating task conflict in multi-task optimization, the
relationship between conflict and negative transfer is still poorly understood.
Quantifying when differences between task gradients are beneficial,
and when they are conflicting, is an open problem; prior work on 
predicting negative transfer between tasks relies on other
heuristics~\citep{pmlr-v119-standley20a,fifty2021efficiently,liuautolambda}.
Therefore, we study how framing multi-task learning in a text-to-text
paradigm affects the relationship between tasks from the perspective of
{\it both} task conflict and negative transfer.

% Despite progress in 
% 
% In practice, however, the multi-task objective can yield a 
% optimization challenges:
% {\it Multi-task conflict} refers to differences in properties of the shared
% parameter gradients $\nabla_\theta \mathcal{L}_k(\cdot)$ across tasks.
% Intuitively, conflict arises in the \emph{directions} between gradients and the
% \emph{magnitudes} across gradients.
% In theory, task conflict in $\theta$ differentiates multi-task and single-task
% learning, by guiding optimization towards regions of the parameter space that
% are difficult to find with a single-task objective.
% However, it is well established that task-conflict can also yield optimization
% difficulties which make jointly minimizing all task objective difficult.
% It is well established that multi-task conflict contributes to negative
% transfer in multi-task learning~\citep{mtsurvey-ruder2017overview}.

\subsection{Substantial Differences Between Canonical and Text-to-Text MTL}
Despite rising interest in text-to-text multi-task learners~\citep{T5-raffel2020exploring,sanh2021multitask}, 
little attention has been given to the effects of text-to-text learning
on multi-task conflict and negative transfer.
We aim to quantify the degree to which different aspects of text-to-text
architectures may mitigate or exacerbate multi-task conflict, and it's
impact on transfer.
We identify three key factors that distinguish text-to-text MTL from canonical
multi-task architectures:
%\squishenum
\begin{description}
\item[F1] The head for each task is an auto-regressive language model,
	often with fewer constraints than task-specific heads.
	All tasks use the same language-modeling loss function
	$\ell_k (f_k(x), y) = \ell_{LM} (f_k(x), r_k(y))$.
% $\mathcal{L}_k(\theta, \beta_k) = \mathcal{L}_{LM}(p_k, r_k, \theta, \beta_k)$.
\item[F2] There are no task-specific parameters $\phi_k$.
	Instead, all parameters $\theta$ are shared across all tasks, 
	and the only gradient necessary for learning is 
	$\nabla_\theta^{MT}$.
\item[F3]
Finally, the MTL objective depends on $p_k$ and $r_k(\cdot)$, fixed natural
language sequences which may or may not encode semantic properties of
task $k$ and it's output space.
\end{description}
%\squishenumend
Factors \fact{1} \& \fact{2} reflect two key architectural decisions about the
MTL model and objective, which may naturally affect multi-task conflict.
For example, removing task specific parameters (Factor \fact{2}) necessitates
specifying tasks at the input; intermediate representations can no longer be
task agnostic, as they now must propagate task information to the decoder.
Alternatively, Factor \fact{3} is concerned with what the model learns during
training; if $p_k$ or $r_k$ are semantically rich, a model may {\it learn} to
leverage them to infer task relations during training.\footnote{Recent
	work has noted that text-to-text learners rarely
	understand the prompts they are given during fine-tuning.
	See \autoref{sec:prompt-understanding} for more discussion.}
The goal of this work is to empirically study how each of these factors
impacts task conflict and negative transfer in MTL.

\section{Factors \fact{1} \& \fact{2}}
\label{sec:fact-1-2}

\begin{table*}[t!]
	\begin{center}
		\scalebox{0.72}{
\begin{tabular} {l l | c c c c c c c c c| c}

\toprule
\multicolumn{11}{c}{\textit{GLUE}} \\
%\midrule
Architecture & Training & CoLA & SST & MSRPC & STSB & QQP & MNLI & MNLI-mm & QNLI & RTE & {\bf Avg} \\

\midrule
% \multicolumn{1}{l}{\textit{T5 - Singletask}} \\
Canonical & Single-Task & 36.20 & 90.29 & 80.80 & 87.46 & 88.69 & 80.34 & 80.72 & 87.96 & 61.25 & 78.06 \\
Canonical & Multi-Task &  30.27 \cellcolor{red!59.25} & 89.98 \cellcolor{red!3.06} & 82.60 \cellcolor{green!17.97} & 87.58 \cellcolor{green!1.26} & 87.89 \cellcolor{red!8.01} & 78.21 \cellcolor{red!21.26} & 78.44 \cellcolor{red!22.75} & 85.84 \cellcolor{red!21.17} & 64.98 \cellcolor{green!37.30} & 76.20 \cellcolor{red!18.59} \\
\midrule
% \multicolumn{1}{l}{\textit{T5 - Multitask}} \\
Text-to-Text & Single-Task & 36.28 & 90.44 & 82.92 & 64.27 & 88.83 & 80.60 & 81.27 & 86.87 & 59.33 & 77.55 \\

Text-to-Text-ID & Multi-Task &  28.64 \cellcolor{red!76.42} & 90.11 \cellcolor{red!3.34} & 84.62 \cellcolor{green!16.95} & 62.08 \cellcolor{red!21.93} & 86.91 \cellcolor{red!19.19} & 77.44 \cellcolor{red!31.58} & 78.14 \cellcolor{red!31.31} & 84.72 \cellcolor{red!21.46} & 65.70 \cellcolor{green!50.78} & 73.15 \cellcolor{red!43.95} \\  

Text-to-Text & Multi-Task &  35.12 \cellcolor{red!11.56} & 90.71 \cellcolor{green!2.68} & 85.21 \cellcolor{green!22.88} & 67.92 \cellcolor{green!36.48} & 87.51 \cellcolor{red!13.26} & 78.98 \cellcolor{red!16.20} & 79.73 \cellcolor{red!15.39} & 86.30 \cellcolor{red!5.64} & 68.47 \cellcolor{green!50.46} & 75.55 \cellcolor{red!19.96} \\

% \midrule
% \multicolumn{1}{l}{\textit{T5 - Singletask}} \\
% \midrule
% Canonical & \\
% Unified & \\
% 
% \midrule
% \multicolumn{1}{l}{\textit{T5 - Multitask}} \\
% \midrule
% Canonical & \\
% Unified & \\

%\bottomrule

\midrule
%\toprule
\multicolumn{11}{c}{\textit{DecaNLP}} \\
%\midrule
Architecture & Training & SST & MNLI & IWSLT & CNN/DM & Seq2SQL & SQUAD & QA-SRL & QA-ZRE & Wino & {\bf Avg} \\

\midrule
% \multicolumn{1}{l}{\textit{T5 - Singletask}} \\
Canonical & Single-Task & 88.74 & 80.40 & 23.55 & 40.22 & 60.28 & 66.57 & 67.34 & 54.85 & 33.93 & 53.41 \\
Canonical & Multi-Task &  90.92 \cellcolor{green!21.78} & 81.15 \cellcolor{green!7.57} & 21.77 \cellcolor{red!17.84} & 39.48 \cellcolor{red!7.46} & 57.25 \cellcolor{red!30.29} & 67.29 \cellcolor{green!7.24} & 70.07 \cellcolor{green!27.25} & 65.79 \cellcolor{green!50.32} & 36.90 \cellcolor{green!29.76} & 58.96 \cellcolor{green!55.48} \\
\midrule

% \multicolumn{1}{l}{\textit{T5 - Multitask}} \\
Text-to-Text & Single-Task & 89.69 & 80.06 & 23.74 & 40.16 & 60.26 & 73.00 & 63.45 & 47.43 & 27.38 & 56.13 \\

Text-to-Text-ID & Multi-Task &  91.61 \cellcolor{green!19.17} & 81.10 \cellcolor{green!10.38} & 23.05 \cellcolor{red!6.93} & 39.88 \cellcolor{red!2.85} & 58.52 \cellcolor{red!17.39} & 72.53 \cellcolor{red!4.72} & 62.90 \cellcolor{red!5.43} & 55.57 \cellcolor{green!50.42} & 31.85 \cellcolor{green!44.64} & 57.44 \cellcolor{green!13.14} \\

Text-to-Text & Multi-Task &  90.33 \cellcolor{green!6.41} & 80.08 \cellcolor{green!0.24} & 22.41 \cellcolor{red!13.28} & 39.54 \cellcolor{red!6.24} & 58.13 \cellcolor{red!21.33} & 72.72 \cellcolor{red!2.83} & 73.85 \cellcolor{green!50.99} & 59.33 \cellcolor{green!50.03} & 47.62 \cellcolor{green!50.38} & 60.45 \cellcolor{green!43.15} \\

\bottomrule
\end{tabular}
}
\end{center}
\caption{\label{table:main}
	Negative transfer between text-to-text and canonical models.
	Results are averaged over 3 random seeds.
	Multi-task model performance is highlighted in red if the performance is lower than single-task performance
	(negative transfer) and green if the performance is higher (positive transfer).
	{\bf We can see that negative (and positive) transfer across tasks are similar
	across text-to-text and canonical MTL, both at a benchmark level and an individual task level.}
}
\end{table*}

\paragraph{Models}
We use T5~\citep{T5-raffel2020exploring}, pre-trained on the C4 language
modeling task, as the base of our models.
A pre-trained T5 model can provide rich representations of inputs which can be
passed to canonical MTL heads~\citep{ni-etal-2022-sentence} or a language model
decoder for text-to-text learning.
Our canonical models simply attach task-specific heads to the outputs
of a shared T5 encoder, and jointly fine-tune all task-specific and shared
parameters.
For more details on the canonical head and loss for each task,
see \autoref{app:task-specific-heads}.

To move from canonical models to models exhibiting Factor \fact{1}, we replace
each task-specific head with a pre-trained T5 decoder head;
that is, for all tasks $k$ we replace
$h_{\phi_k}:\mathbb{R}^d \rightarrow \mathcal{Y}_k$ with
$j_{\phi_k}:\mathbb{R}^d \rightarrow \mathcal{X}$.
Note that each decoder is still parameterized with task-specific parameters
$\phi_k$, and therefore each task is still specified by which decoder is used
at test-time.
We refer to this architecture as \emph{Text-to-Text-ID}: models which treat all
tasks as text-to-text problems, but which still leverage independent heads for
task specification.\footnote{
% Factor \fact{1} removes task-specific constraints from the output space of the model: rather than only
% predicting into a space of $\mathbb{R}^2$ for binary classification, the model must predict
% into sequences of tokens in $\mathcal{X}$.
We explore the effects of re-initializing the decoder parameters of text-to-text-ID models in \autoref{app:no-pretraining-decoder}.
}

Factor \fact{2} requires a unified output space for each task in order to remove
task-specific parameters.
% \footnote{
% For a subset of the tasks we consider, it is possible to construct a joint
% output space by considering the cartesian product of each task-specific output
% space, rather than moving the output space to natural language.
% However, this quickly becomes infeasible for tasks which have }
To test the effects of Factor \fact{2} on MTL, we
start from Factor \fact{1} models (text-to-text-ID)
and combine all task heads into a single decoder head
shared across all tasks, $j_{\theta_d}$.
This yields a fully unified \emph{Text-to-Text} architecture, equivalent
to the original T5 MTL setting~\citep{T5-raffel2020exploring}.
Factor \fact{2} removes task-specific parameters from the model, and therefore
we must specify tasks at the input.
For the following experiments we utilize the default prompts and
output spaces listed in \autoref{app:data};
in \autoref{sec:prompt-understanding} we explore how Factor \fact{3} (different
task prompts and output spaces) affect our results.

\paragraph{Datasets}
We study negative transfer and multi-task conflict on two common
NLP multi-task benchmarks.
The first is GLUE~\citep{wang-etal-2018-glue}, a set of 8
natural language understanding tasks.The tasks in GLUE can be separated into 7
classification tasks and one regression task;
thus, canonical task heads consist of a single, linear layer that projects
representations into 1, 2, or 3 dimensions.
% T5 has also demonstrated impressive performance on GLUE when fine-tuned in a
% text-to-text manner, rather than with classification or regression heads.
The second dataset is the NLP Decathlon~\citep[DecaNLP;][]{McCann2018decaNLP}
benchmark, which frames 9 NLP tasks as Q\&A problems.
DecaNLP consists of span-labeling, classification,
and text-to-text tasks; canonical heads for DecaNLP include linear
classification layers, span-labeling classifiers,
and T5 decoders for text-to-text tasks.
DecaNLP therefore represents a setting with more diverse 
canonical heads than GLUE, making it an interesting case study against
GLUE.
\citet{T5-raffel2020exploring} and \citet{McCann2018decaNLP} provide default
$p_k$ and $r_k$ for GLUE and DecaNLP respectively, which we use in this section;
for more details about the individual tasks of each benchmark and their default
prompts see \autoref{app:data}.\footnote{
We exclude 2 tasks from our study. The WNLI test set in GLUE
is notoriously problematic for models of the size we consider, and we
exclude it to avoid confusion over negative transfer.
We additionally exclude the WoZ dataset in DecaNLP, because the canonical 
output space of WoZ is not suitable for the single-encoder canonical setup
that we rely on.}

\subsection{Negative Transfer is Similar Across Factors \fact{1} \& \fact{2}}
\label{sec:empirical-neg-transfer}

We compare single- and multi-task models, over both canonical and text-to-text
settings, in \autoref{table:main}.
Text-to-Text-ID models have independent heads, representing Factor \fact{1},
and Text-to-Text multi-task models represent Factors \fact{1} \& \fact{2}.
For both GLUE and DecaNLP models,
canonical and text-to-text models perform comparably on most tasks.
For GLUE, text-to-text models perform marginally below canonical models, on average,
notably due to STS-B, a regression task which suffers from the usage of a fixed vocabulary
for numeric outputs.
For DecaNLP, text-to-text models slightly outperform canonical models: 7 out of
the 9 tasks are text-to-text or span-labeling tasks, the latter of which may be more amenable
to text-to-text learning.

Our results illustrate that both text-to-text and canonical models exhibit
similar amounts of negative and positive transfer.
On GLUE, MTL reduces performance by nearly 2.5\% on average in \emph{both}
text-to-text and canonical settings.
In DecaNLP, MTL generally exhibits
\emph{positive transfer}---a trend also noted in the original
paper~\citep{McCann2018decaNLP}---suggesting that these tasks benefit each other.
However, again canonical and text-to-text exhibit similar \emph{levels} of
positive transfer (a 8-10\% increase in performance).
Moreover, the similarities continue at the individual task-level: tasks which
experience positive or negative transfer in the canonical models often
experience the same amount of positive or negative transfer in text-to-text
models.

We additionally see that Text-to-Text-ID models, which add Factor \fact{1} to canonical models,
are generally worse than Text-to-Text models, which combine Factors \fact{1} and
\fact{2}.
In tasks where this difference is significant, such as STS-B or CoLA,
we identify this as an issue with the pre-trained decoder (\autoref{app:no-pretraining-decoder});
re-initializing the decoders boosts the strength of STS-B and CoLA, suggesting an
incompatibility between language model pre-training and regression-like tasks.
For other tasks, the effect of Factor \fact{2} is less significant: 
tasks such as translation appear to benefit slightly from task-specific
decoders, whereas other tasks, e.g. span-labeling tasks, benefit from a shared decoder.
On average, Text-to-Text models outperform Text-to-Text-ID models significantly,
suggesting that the benefits of Factor \fact{2} for some tasks outweigh the
costs to other tasks in terms of performance, on the settings we consider.
In short, when framing multi-task learning as a text-to-text problem
(Factor \fact{1}), it is typically better to share the decoder (Factor \fact{2}).

Despite what could be considered radically different approaches to MTL, our
findings suggest that Factors \fact{1} and \fact{2} may not have a significant
effect on the optimization landscape of multi-task learning.
Factors \fact{1} and \fact{2} combined frame MTL as a multi-domain
problem, where tasks represent distributions over the same problem
space ($\mathcal{X} \rightarrow \mathcal{X}$).
Our results suggest that this re-framing is not sufficient, alone, to mitigate
negative transfer.\footnote{We corroborate our results from this section on GLUE using a larger model 
(T5-Base) in \autoref{app:larger-model}.}

\section{Analyzing Multi-Task Conflict}

\begin{figure*}[t!]
    \centering
    % \begin{subfigure}[t]{0.49\textwidth}
    % \caption{MQAN}
    % \includegraphics[scale=0.09,trim=40px 0 0 0]{MQAN-architectures.pdf}
    % \end{subfigure}
    \hfill
    \begin{subfigure}[t]{.49\textwidth}
    \caption{GLUE}
    \includegraphics[scale=0.28,trim=40px 80px 0 70px]{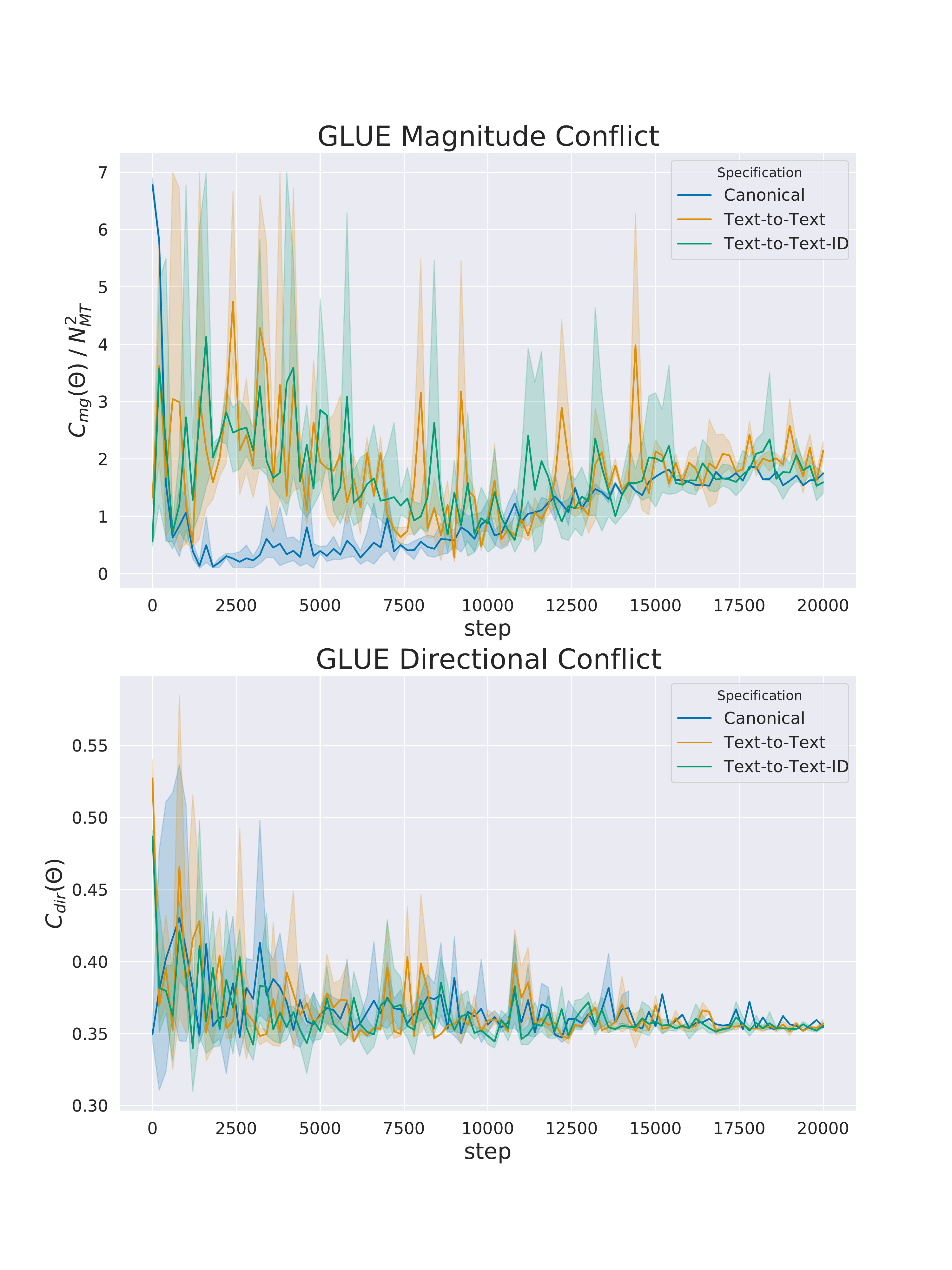}
    \end{subfigure}
    \begin{subfigure}[t]{.49\textwidth}
    \caption{DecaNLP}
    \includegraphics[scale=0.28,trim=40px 80px 0 70px]{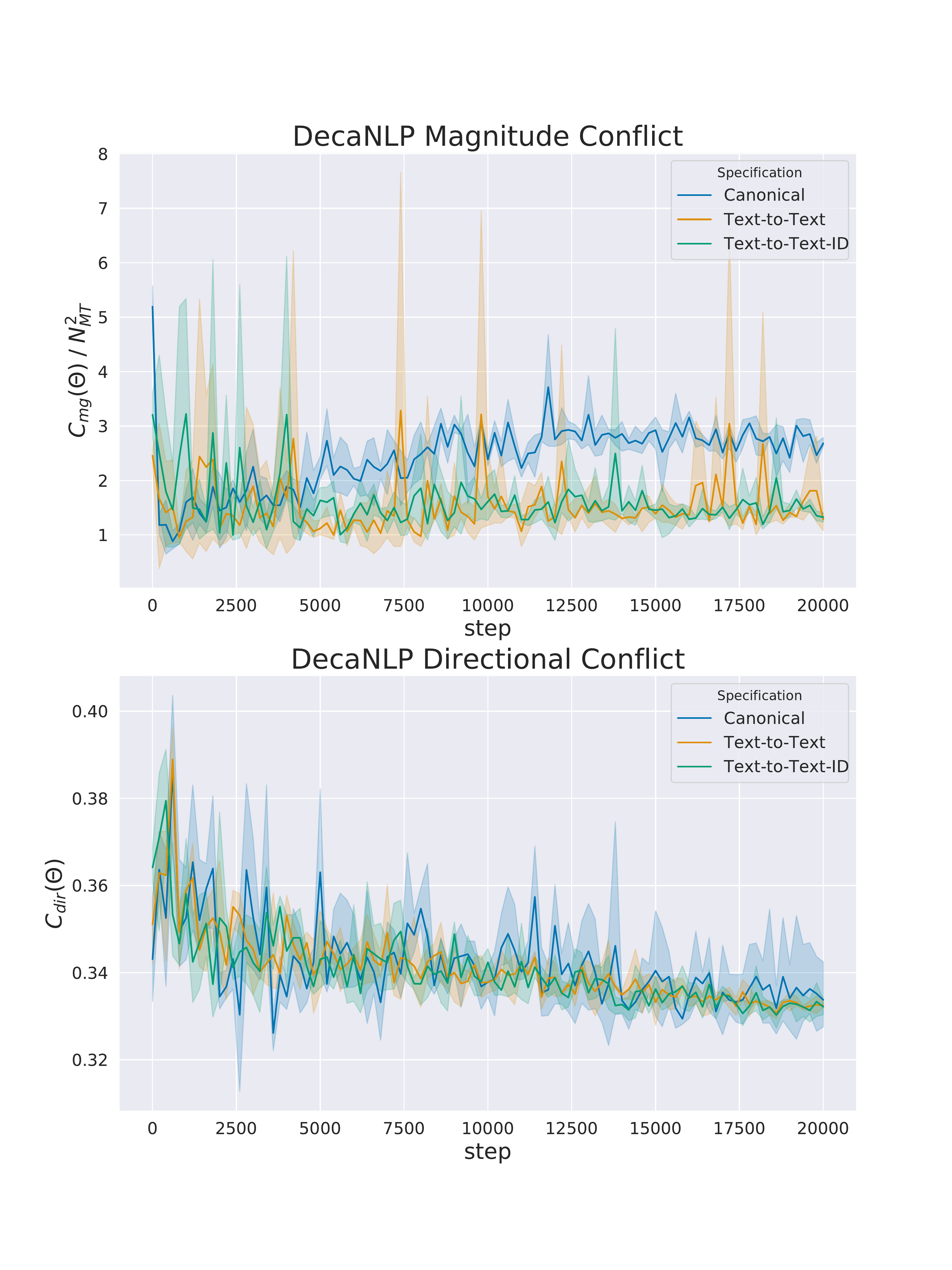}
    \end{subfigure}
	\caption{\label{fig:conflict-graphs} Magnitude conflict ($C_{mg}$, top, normalized by average
	gradient magnitude) and
	directional conflict ($C_{dir}$, bottom) across GLUE (left) and DecaNLP (right)
	tasks in canonical, text-to-text, and text-to-text-ID models.
	Directional conflict in the encoder follows similar trajectories across different architectures,
	initially noisy and eventually stabilizing to similar values; however,
	magnitude conflict differs across settings, and is likely tied to the homogeneity of
	canonical task loss functions.}
\end{figure*}

Our results in \autoref{sec:empirical-neg-transfer} suggest that negative transfer
is not impacted by framing MTL as a text-to-text problem.
We next empirically examine whether this is borne out via \emph{multi-task conflict}.
Multi-task conflict is typically thought to be a key factor in
negative transfer (\autoref{sec:mtl-conflict}).
Multi-task conflict can be summarized as the differences between individual task gradients;
while some amount of conflict is necessary for multi-task learning to benefit over single-task
learning, too much conflict can create optimization difficulties.
Given that we see equivalent amounts of negative and positive transfer across canonical
and text-to-text models, we expect to see similar levels of conflict as well.

\paragraph{Measuring Conflict}
We measure two types of conflict in this work: magnitude and direction.
Let $\Theta$ be all the parameters of a model, $\nabla_\theta \mathcal{L}_k(\cdot)$ be
the gradient of $\theta$ for task $k$ with parameters $\Theta$, and $\nabla_\theta^{MT}$
be the \emph{multi-task} gradient of $\theta$ with total parameters $\Theta$ (as in 
\autoref{sec:background}).
We measure \emph{magnitude conflict}, $C_{mg}$, as the variance across each
task's gradient magnitude:
\begin{equation*}
	C_{mg}(\Theta) = \frac{1}{K}\sum_{k=1}^K \bigg(\big|\big|\nabla_\theta\mathcal{L}_k(\cdot)
\big|\big|_2^2
- \big|\big|\nabla_\theta^{MT}\big|\big|_2^2
\bigg)^2
\end{equation*}
A high value of $C_{mg}$ indicates high conflict between the magnitudes of
task gradients, which may lead to optimization being dominated by a few
high-magnitude tasks.
$C_{mg}$ is highly correlated with the magnitude of the loss at each step and the
average gradient magnitude.
To control for text-to-text models having training losses several orders of magnitude
higher than classification losses at the beginning of training,
we normalize the variance of gradient norms $C_{mg}$ by the average gradient norm
squared $N_{MT}^2$ at each step.
Thus, when comparing gradient conflict across different models, the resulting metric
measures the gradient variance if each model had the same \emph{average}
gradient norm.

We measure {\it directional conflict}, $C_{dir}$, as the norm of the multi-task
gradient after normalizing $\nabla_\theta$ to magnitude $1$:
\begin{equation*}
	C_{dir}(\Theta) = \bigg|\bigg| \frac{1}{K} \sum_{k=1}^K \frac{\nabla_\theta \mathcal{L}_k(\cdot)}{
	||\nabla_\theta \mathcal{L}_k(\cdot) ||_2} 
	\bigg|\bigg|_2^2
	- \frac{1}{K}
\end{equation*}
This metric is equivalent, up to a constant, to the average pairwise cosine
similarity between all task gradients, which is a standard measurement for
directional conflict in multi-task
optimization~\citep{pcgrad-yu2020gradient,gradvaccine-wang2020gradient}.
% \footnote{The constant, simplified, is $\frac{K}{K-1}$,
% where $K$ is the number of tasks. We measure cosine similarity in this way as
% it requires maintaining a single running summation during learning, and is thus
% much more efficient to compute than pairwise similarity metrics.}

\paragraph{Comparing conflict}
Our goal is to compare how task conflict changes across canonical and text-to-text models,
and specifically, how Factors \fact{1} and \fact{2} affect task conflict.
To do this, we compare $C_{dir}$ and $C_{mg}$ in Canonical, Text-to-Text-ID, and Text-to-Text
models; in our comparisons across these architectures we measure conflict in
\emph{the encoder only}, which ensures that we are measuring task-conflict across not only the
same number of parameters, but over the same architecture as well (a T5 encoder).

\paragraph{Results}

We plot $C_{mg}$ and $C_{dir}$ across the training trajectories of canonical, text-to-text-ID, and text-to-text
models in \autoref{fig:conflict-graphs}.
We see that \emph{directional} conflict remains similar across all settings. That is, models begin training
with relatively noisy average cosine similarities ,
and gradually over time conflict increases (pairwise cosine similarity decreases)
and becomes more stable.
Throughout training, text-to-text and canonical models maintain similar values of $C_{dir}$, and
have similar variances in both GLUE and DecaNLP settings.
This result corroborates our findings in \autoref{sec:empirical-neg-transfer}: multi-task conflict,
in the encoder parameters, remains consistent regardless of how tasks are specified, and this translates
to similar levels of negative transfer in test-set performance.

\begin{table*}[t!]
	\begin{center}
		\scalebox{0.8}{
\begin{tabular} {l | c c c c c c c c c| c}

\toprule

Prompt & CoLA & SST & MSRPC & STSB & QQP & MNLI & MNLI-mm & QNLI & RTE & {\bf Sum} \\
\midrule
Single-Task (Null Prompt) & 36.28 & 90.44 & 82.92 & 64.27 & 88.83 & 80.60 & 81.27 & 86.87 & 59.33 & 77.55 \\
\midrule
Null Prompt &  29.56 \cellcolor{red!50.18} & 90.34 \cellcolor{red!1.05} & 83.46 \cellcolor{green!5.31} & 48.20 \cellcolor{red!50.73} & 86.88 \cellcolor{red!19.52} & 76.71 \cellcolor{red!38.87} & 76.40 \cellcolor{red!48.70} & 85.08 \cellcolor{red!17.89} & 50.99 \cellcolor{red!50.33} & 50.99 \cellcolor{red!50.54} \\
Default Prompt &  35.12 \cellcolor{red!11.56} & 90.71 \cellcolor{green!2.68} & 85.21 \cellcolor{green!22.88} & 67.92 \cellcolor{green!36.48} & 87.51 \cellcolor{red!13.26} & 78.98 \cellcolor{red!16.20} & 79.73 \cellcolor{red!15.39} & 86.30 \cellcolor{red!5.64} & 68.47 \cellcolor{green!50.46} & 75.55 \cellcolor{red!19.96} \\
Diverse Prompt &  32.39 \cellcolor{red!38.90} & 89.94 \cellcolor{red!5.06} & 79.96 \cellcolor{red!29.62} & 66.30 \cellcolor{green!20.23} & 87.31 \cellcolor{red!15.27} & 77.22 \cellcolor{red!33.77} & 77.87 \cellcolor{red!34.01} & 86.25 \cellcolor{red!6.18} & 65.61 \cellcolor{green!50.88} & 73.65 \cellcolor{red!38.97} \\
\bottomrule
\end{tabular}
}
\end{center}
\caption{\label{table:prompts}
	Multi-task model performance for GLUE models trained on null, default, and diverse prompts.
	All results are averaged over 3 random seeds.
	While diverse prompts may help with zero-shot transfer~\citet{sanh2021multitask},
	they do not necessarily improve negative transfer in text-to-text multi-task models.}
\end{table*}

Magnitude conflict, however, exhibits different trends across each setting.
On GLUE, canonical models have lower magnitude conflict that text-to-text models early into training,
but this effect diminishes over time; eventually text-to-text and canonical models converge
to similar levels of conflict.
DecaNLP models exhibit a different effect: canonical models actually exhibit higher magnitude conflict
than text-to-text models, which is exacerbated as training progresses.
This result may be partially explained by the fact that, in GLUE, 7 out of 8 tasks share the same loss
function (cross-entropy for classification), leading to more consistent magnitudes in loss towards the
beginning of training; in contrast, DecaNLP canonical models compute span-labeling losses, 
classification losses, and conditional language model losses.
Notably, however, the differences in \emph{gradient} magnitude do not seem to have any bearing on
negative transfer: canonical models have lower magnitude conflict in GLUE and higher magnitude conflict in
DecaNLP than text-to-text models, yet both benchmarks exhibit similar transfer across canonical and text-to-text models.

Finally, we note that conflict in the \emph{encoder} is largely consistent between Text-to-Text and
Text-to-Text-ID architectures.
Therefore, we conjecture that, if multi-task conflict is beneficial when moving from independent heads
to a single joint head, it is largely due to conflict that occurs in the decoder parameters.

Our findings suggest that prior work in multi-task learning which studies canonical models,
especially work focused on multi-task conflict and negative transfer (\autoref{sec:mtl-conflict}),
may still be relevant in text-to-text MTL.
We observe that shifting from canonical to text-to-text MTL has little effect on both negative transfer
and multi-task conflict.
The fully-shared parameterization of
text-to-text models does not translate into reductions in task conflict, nor major improvements in
positive transfer between tasks.
Future work on improving multi-task optimization by addressing conflict in the text-to-text setting
is therefore a promising avenue.

\section{Factor \fact{3}: Task Prompts \& Labels}
\label{sec:prompt-understanding}

We have seen that Factors \fact{1} \& \fact{2} have little effect
on transfer and task conflict, in our settings.
We now turn to Factor \fact{3}: the form of $p_k$ and $r_k$.
In text-to-text MTL, tasks are specified via natural language prompts
which can range in semantic richness from nondescript tokens, such as ``SST: '',
to descriptions of the task:
``Is the sentiment of this review positive or negative?: ''.
In theory, a T2T multi-task model may leverage the semantics of
task descriptions to implicitly align related tasks during learning, 
especially when leveraging a pre-trained model which
exhibits non-trivial prompt understanding.
However, \citet{promptfail-logan2021cutting} demonstrated that T5 could be fine-tuned with null prompts and retain
competitive accuracy, suggesting that models prefer to memorize
prompts regardless of their semantic information.
If T2T multi-task models memorize task prompts, the form of
$p_k$ and $r_k$ should have no impact on task conflict.

\subsection{Leveraging Diverse Prompts}
\label{sec:diverse-prompt}

Recently, \citet{sanh2021multitask} proposed T0, a T5 architecture
fine-tuned in a massively multi-task manner with a \emph{diverse} set
of crowd-sourced prompts for each task: specifically, they leverage crowd-sourced
workers to generate multiple distinct, semantically
informative $p_k$ and $r_k$ {\it for each task}.
\citet{sanh2021multitask} find that a model trained with these diverse, descriptive
prompts have significantly stronger {\it zero-shot} capabilities than models trained
on one prompt per task, suggesting T0 has learned to leverage semantic information
in task prompts.
This result has been called into question by \citet{promptfail-webson2021promptbased}, who
demonstrate that few-shot performance, even with T0, can still be competitive when using
nonsense prompts; however, the strength of T0's zero-shot performance suggests that
diversifying the task prompts during learning may improve the model's understanding
of task descriptions {\it to some extent}:
does this correlate with less negative transfer when fine-tuning?

To test this, we leverage \texttt{promptsource}, a repository of diverse prompts for a large set of NLP
tasks for text-to-text learning~\citep{sanh2021multitask}.
For each task $k$ in GLUE, \texttt{promptsource} contains 4-7 pairs of prompts and text-labels $(p_k, r_k)$.
We re-train our T2T multi-task GLUE models using a randomly sampled pair ($p_k$, $r_k$) for each sample in task $k$;
We term this model the \emph{diverse} prompt model, and compare it's multi-task performance to
the \emph{default} prompt model, a model which uses task-specific
tokens---similar to those in the original T5 approach---which encode no semantic information.
Additionally, we train a text-to-text multi-task model with \emph{null} prompts, e.g.
no prompts. This model must learn which task it is performing by the distribution
of the input samples alone.

\begin{figure}[t!]
    \centering
    % \begin{subfigure}[t]{0.49\textwidth}
    % \caption{MQAN}
    % \includegraphics[scale=0.09,trim=40px 0 0 0]{MQAN-architectures.pdf}
    % \end{subfigure}
    \hfill
    \begin{subfigure}[t]{.49\textwidth}
    % \caption{GLUE}
    \includegraphics[scale=0.28,trim=40px 80px 0 70px]{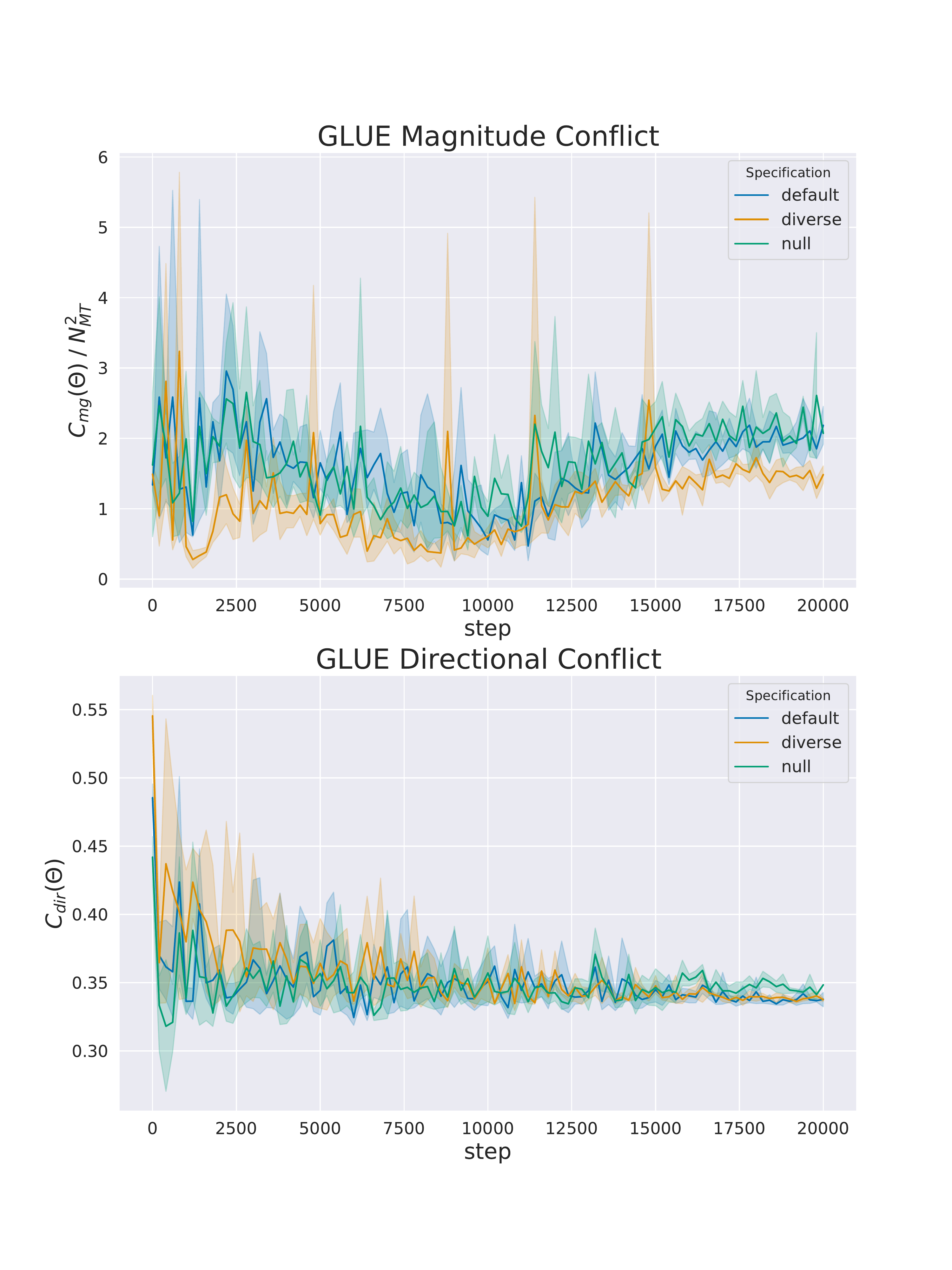}
    \end{subfigure}
    % \begin{subfigure}[t]{.49\textwidth}
    % \caption{DecaNLP}
    % \includegraphics[scale=0.28,trim=40px 80px 0 70px]{deca-neg-transfer.pdf}
    % \end{subfigure}
	\caption{\label{fig:prompt-conflict-graphs} Magnitude conflict ($C_{mg}$, top) and
	directional conflict ($C_{dir}$, bottom) in GLUE models using null, default, and
	diverse prompts.
	We find that increasing prompt diversity generally \emph{lowers} task conflict,
	suggesting that it pushes task distributions closer together.
	}
\end{figure}

We plot the multi-task results of null, default, and diverse T2T models in \autoref{table:prompts}, as well
as single-task text-to-text performance; we plot the multi-task conflict of models trained with
the 3 different prompts in \autoref{fig:prompt-conflict-graphs}.
The cells of \autoref{table:prompts} are highlighted by their magnitude of positive (green) or negative (red)
transfer from single-task T2T models, which are trained with null prompts.
We find that null prompts perform poorly in MTL, suggesting that T2T
models struggle to determine task specification based on the input text distribution alone.
Default prompts, used by the original T5 paper, which are fixed tokens pre-pended to each input,
perform much better than null prompts.
This result demonstrates the importance of specifying the task at the input during text-to-text MTL; while
single-task models can be trained with null, and even nonsensical, prompts to competitive accuracy~\citep{promptfail-logan2021cutting},
MTL requires the text distribution to differentiate the tasks explicitly.

We additionally find that diverse prompts increase negative transfer compared to default prompts,
suggesting the model is failing to 
leverage the information in $p_k$ during training.
We note that a limitation of our work is the number of tasks we consider; it is possible that,
under a MTL regime with 9 tasks, even a diverse set of prompts is not enough to improve prompt understanding.
As a result, it may be that introducing multiple prompts per task serves only to confuse the
model as to how tasks are specified, lowering test performance.

Finally, in \autoref{fig:prompt-conflict-graphs} we see that, once again, directional conflict
is largely similar across training paradigms.
Null prompt models see slightly less directional conflict towards the end of training,
potentially due to the fact that, without task-specifications, task distributions overlap;
however, lower directional conflict is {\it not} correlated with stronger task performance
in this case.
Additionally, diverse prompts appear to have consistently lower magnitude conflict
throughout training; again, lower conflict is not indicative of a stronger multi-task
model, as default prompt models perform the best out of the 3 specifications.

\subsection{Overlap in the Output Space}

\begin{figure}[t!]
    \centering
    % \begin{subfigure}[t]{0.49\textwidth}
    % \caption{MQAN}
    % \includegraphics[scale=0.09,trim=40px 0 0 0]{MQAN-architectures.pdf}
    % \end{subfigure}
    \hfill
    \begin{subfigure}[t]{.49\textwidth}
    % \caption{GLUE}
\includegraphics[scale=0.28,trim=40px 50px 0 0]{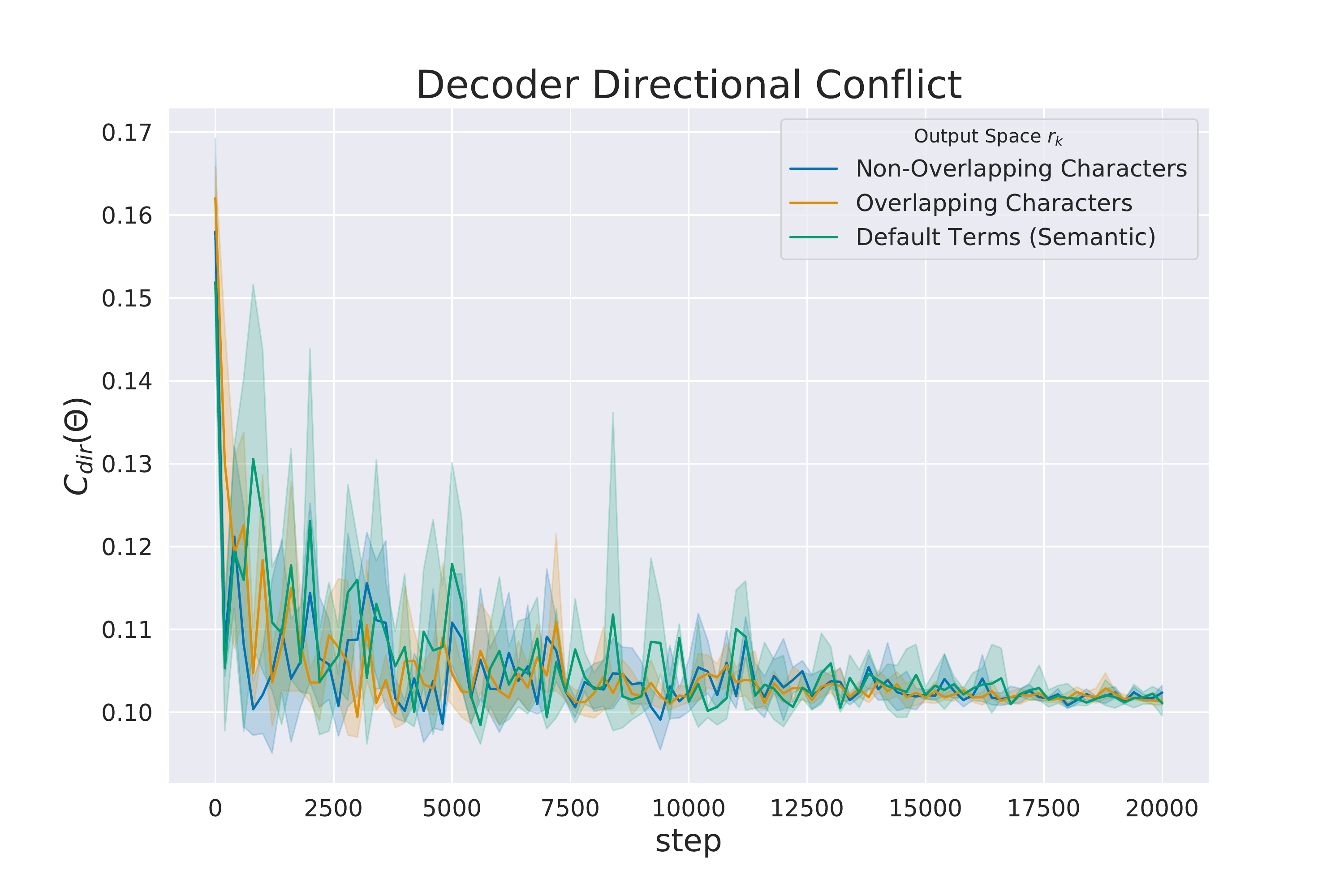}
    \end{subfigure}
        \hfill
    \begin{subfigure}[t]{.49\textwidth}
    % \caption{GLUE}
    \includegraphics[scale=0.28,trim=40px 50px 0 0]{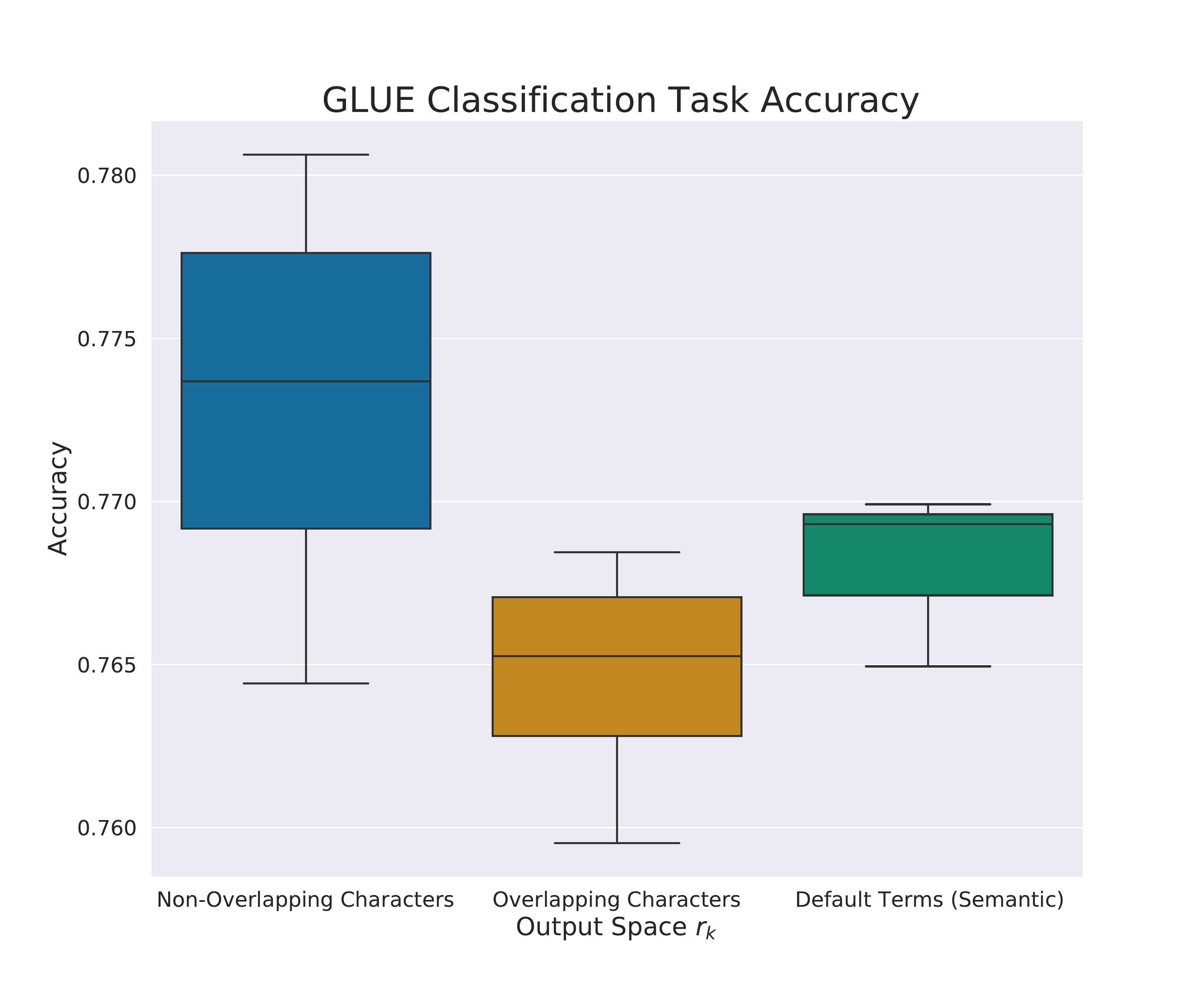}
    \end{subfigure}
    % \begin{subfigure}[t]{.49\textwidth}
    % \caption{DecaNLP}
    % \includegraphics[scale=0.28,trim=40px 80px 0 70px]{deca-neg-transfer.pdf}
    % \end{subfigure}
	\caption{\label{fig:output-conflict-graphs} (Top):
	Directional conflict ($C_{dir}$) in the {\it decoder} of GLUE
	models when using different classification output spaces.
	The output space used, and the overlap across tasks,
	has no observable effect on multi-task conflict.
	(Bottom): The impact of classification output spaces on multi-task
	performance. Overlap in the output space appears to have an effect 
	on model accuracy: output spaces with less overlap perform better.}
\end{figure}

\autoref{sec:diverse-prompt} suggests that, in the settings we
study, the semantics of task specifications
have little effect on MTL.
To that end, we test a much simpler artifact of Factor \fact{3}: overlap
across the output spaces of classification tasks.
If the semantics of $r_k$ is not important to multi-task performance, then 
hypothetically the selection of terms for class labels should not matter (e.g.
using the letter ``z'' versus the term ``entailment'').
However, one factor that {\it may} impact multi-task learning is the amount of
overlap in class label terms across different tasks; is it easier to learn
multiple classification tasks if their label terms are distinct?
To study this, we create two additional $r_k$ for each classification task in
GLUE:
the first is {\it non-overlapping characters}, where we sample a random
character for each class label, ensuring that no character is used twice
across all tasks.
The second is {\it overlapping characters}, where we instead sample a few
characters, which are re-used for class labels over all tasks.
Finally, we consider the {\it default terms} for class labels, which are
semantically meaningful, multi-token terms.

We plot conflict in the {\it decoder} of the model and average
classification task accuracy, in \autoref{fig:output-conflict-graphs}.
We find that, similar to other factors that we study in this work, different
output spaces have no observable impact on task conflict (magnitude
conflict exhibits the same trend).
However, we {\it do} observe impacts on model accuracy:
models with less overlap in their output space tend to perform better than
models with more overlap,
suggesting it is harmful to re-use class labels across different tasks.\footnote{
Although we emphasize that this may only be the case when the semantics of the
class tokens are not important, i.e. in the few-task regime}

Overall, we find that it is important to specify tasks
in T2T MTL: expecting the model to learn task specification
from the input distribution alone leads to high negative transfer.
However, when the intent is to train a SotA multi-task model via fine-tuning,
these prompts and outputs need not be semantically rich:
simple task-specific tokens often suffice.
Indeed, when MTL is being done in a setting that is \emph{not} massively
multi-task~\citep{sanh2021multitask,aribandi2022ext}, rich and diverse prompts may
actually hurt multi-task learning, rather than benefit it.
Additionally, throughout this work, our findings have highlighted the
disconnect between standard notions of task conflict and negative transfer:
task conflict, particularly across gradient magnitudes, is rarely predictive
of negative transfer in the settings we study.

\section{Conclusion}

\paragraph{Findings} General purpose text-to-text architectures, typically pre-trained on large amounts of data, are increasingly popular both as a topic of academic study and in commercial NLP applications.
By using natural language prompts rather than task-specific parameters, models like T5 can
encode multiple tasks in the same framework with fully shared parameters. Empirically, text-to-text
models often achieve comparable or superior performance compared to task-specific models.
Surprisingly, we find similar task conflict in both canonical and text-to-text models, with similar negative transfer. This begs the question: are text-to-text models fundamentally better multi-task learners than canonical MTL models?
Or is their success largely attributable to other factors, such as increased model capacity and pre-training on more data?
Our findings suggest the latter: text-to-text architectures are \textbf{not} inherently better 
multi-task learners than canonical models.
In particular, when controlling for other factors, we have found that T5 exhibits 
comparable transfer characteristics to canonical MTL
parameterizations, both in terms of performance on held-out data and over the course of optimization by measuring rates of gradient conflict.
To our knowledge, we are the first to directly compare these two distinct learning paradigms on multi-task optimization and transfer.

\paragraph{Future directions} Recently, a series of optimization methods have been proposed to
address the challenges posed by multi-task, or more generally, multi-objective learning~\citep{gradvaccine-wang2020gradient,javaloy2022rotograd}.
If the optimization landscape of multi-task text-to-text models poses similar challenges as canonical MTL parameterizations,  applying special-purpose MTL optimization methods with text-to-text models may result in faster convergence, improvements in performance, or both.
However, more research is needed both to understand the connection between task conflict and negative
transfer, and to develop new optimization techniques better
suited to the fully-shared parameterization used by models like T5.
As such, our results suggest several promising 
avenues for future work, which to our knowledge have lacked clear motivation until now.

\section{Limitations}

A foremost limitation of our work is that we consider only a single  language, English, largely
due to the access of both strong text-to-text models for English as well as
several English multi-task benchmarks.
While we expect that our findings regarding multi-task learning are generalizable
across languages, further experiments involving multi-task benchmarks and pre-trained encoders in other languages are necessary to fully support this claim.

Our work is additionally limited by the multi-task settings we examine.
While we study two popular NLP MTL benchmarks, GLUE and DecaNLP,
both have limitations and prior work has shown wide variance in how well MTL works on different datasets.
Notably, the canonical output space of all tasks is fairly limited, e.g.
our work overall considers regression, classification, span-labeling, and sequence
output spaces.
However, other classical NLP output spaces are also worth studying, such as
constrained structured prediction in Named Entity Recognition or Syntax Parsing.
Finally, our MTL settings consider only 9 tasks to be learned jointly.
A smaller universe of tasks enables us to fairly compare single- and multi-task
models across several different architectures. Recent work has begun
to consider \emph{massively multi-task models} on 100+ tasks simultaneously~\cite{aribandi2021ext5}.
While we expect that our findings will generalize to these extreme settings,
expanding our experiments to larger universes of tasks would validate this claim.\footnote{On the other hand, the present study is easily reproducible on commodity compute
resources.
Code for all of our experiments can be found at \url{https://github.com/davidandym/task-conflict-in-text-to-text-learners}.}

\section*{Acknowledgements}
We would like to thank the anonymous reviewers for the helpful comments and feedback on this work.
This work was supported, in part, by the Human Language Technology Center of Excellence
at Johns Hopkins University.

\bibliography{anthology,custom}
\bibliographystyle{acl_natbib}

%\pagebreak
\appendix

% \begin{figure}[t!]
%     \centering
%     % \begin{subfigure}[t]{0.49\textwidth}
%     % \caption{MQAN}
%     % \includegraphics[scale=0.09,trim=40px 0 0 0]{MQAN-architectures.pdf}
%     % \end{subfigure}
%     \hfill
%     \begin{subfigure}[t]{.49\textwidth}
%     % \caption{T5}
%     \includegraphics[scale=0.1,trim=0 50px 0 0]{Multi-task architectures.pdf}
%     \end{subfigure}
% 	\caption{\label{fig:app-architectures} \emph{Canonical} (top) vs. \emph{Text-to-Text} (bottom)
% 		multi-task T5 architectures.
% 		In this work we are interested in how \emph{negative transfer} and \emph{multi-task conflict}
% 		are affected by moving from Canonical MTL to Text-to-Text MTL in NLP.}
% \end{figure}

\section{Model and Training Details}
\label{app:task-specific-heads}

All of our models start from the \texttt{google/t5-v1\_1-small} checkpoint in the
Huggingface pre-trained model hub.\footnote{\url{https://huggingface.co/google/t5-v1_1-small}}
The \texttt{t5-v1\_1} model parameters are pre-trained only on C4 with the language-modeling objective,
and have not been pre-trained on any multi-task data from GLUE or other datasets.
It also is pre-trained with a slightly more advanced pre-training regime than the original
T5 model -- it uses GEGLU activation, rather than ReLU, and was pre-trained with no dropout.

Our multi-task models are fine-tuned with a learning rate of 5e-4 and a batch-size of $16$, and our single-task models are fine-tuned with a learning rate of 3e-4 and a
batch-size of $16$.
All models use a dropout rate of $0.0$ and gradient clip value of $5$, which we found to empirically work the best with our datasets across all settings.
We fine-tune both single- and multi-task models for a total of $20,000$ steps, and use
the final model for test-set predictions.
We use a held-out validation set, often split from the original training data, to select
all hyper-parameters mentioned above.

\subsection{Canonical Architecture Decoders}

Below we discuss the various task-specific heads and loss functions used for 
canonical models.

\paragraph{Regression Tasks}
For regression tasks we use a simple linear layer which projects sentence
level representations into $\mathbb{R}$.
Following \citep{ni-etal-2022-sentence}, to compute sentence representations from
T5 token encodings, we aggregate all token representations using mean pooling.
For regression tasks the loss function $\mathcal{L}_k$ is mean squared error (MSE).

\paragraph{Classification Tasks}
For classification tasks we use a simple linear layer which projects sentence
level representations into $\mathbb{R}^c$, where $c$ is the number of classes.
Following \citep{ni-etal-2022-sentence}, to compute sentence representations from
T5 token encodings, we aggregate all token representations using mean pooling.
For regression tasks the loss function $\mathcal{L}_k$ is Cross-Entropy.

\paragraph{Span Labeling Tasks}
For span-labeling tasks we use 2 simple linear layers which project token-level
representations into $\mathbb{R}$ -- one layer predicts the start token, and the other
predicts the end token.
We aggregate token-level scores from each classifier into a single vector, and compute
a softmax distribution over that vector to obtain token-level probabilities for start
and end tokens.
The loss function $\mathcal{L}_k$ is the average cross entropy for start and end
predictions.
If the task has the possibility of no answer being in the context, we additionally add a
sentence-level linear binary classifier which predicts whether the context contains the
answer or not, and adds an additional cross-entropy loss to the loss function.

\paragraph{Generative Tasks}
For tasks whose output space is $\mathcal{X}$, e.g. all sequences of a fixed vocabulary,
we use a pre-trained $T5$ decoder.
The $T5$ decoder is an autoregressive decoder, pre-trained on the C4 language modelling
task.
The loss function for these tasks is the same loss function as the pre-training task --
$\mathcal{L}_{LM}$, the average token cross-entropy of the auto-regressive outputs.

\section{Data}
\label{app:data}

\subsection{GLUE}

The GLUE dataset~\citep{wang-etal-2018-glue} consists of 9 tasks - 8 classification
tasks and 1 regression task.
In our experiments we exclude the WNLI dataset -- we found that our models tended to
overfit to the training data and perform below random chance on the test
data.\footnote{See \url{https://gluebenchmark.com/faq} for more on WNLI performance.}
This leaves 7 classification tasks and 1 regression task, which we detail below:
For all datasets, we take $5\%$ of the training dataset as a held out validation
set for hyper-parameter tuning, and use the official validation sets as our held-out
test sets.

\paragraph{CoLA} is a dataset of English acceptability
judgements~\citep{warstadt-etal-2019-neural} - each example consists of a sequence of
words and is annotated with wither or not the sentence is grammatical English.
In GLUE, this task is treated as a \emph{binary classification} task, with sentences
classified as either acceptable or unacceptable.
Our default prompt for this task, following T5, is $p_k =$ ``\emph{cola sentence:}''
and $r_k =$ ``\emph{acceptable}'' for  acceptable sentences or ``\emph{unacceptable}''
otherwise.
The evaluation metric for this task is the Mathews correlation coefficient.

\paragraph{SST} is a dataset of movie reviews and human annotations of their
sentiment~\citep{sst-socher-etal-2013-recursive}.
The dataset is treated as a \emph{binary classification} dataset, with sentences
classified as either positive or negative.
The base prompt $p_k$ = ``\emph{sst2 sentence:}'' and $r_k$ = ``\emph{positive}'' if
the class is positive or ``\emph{negative}'' if the class is negative.
The evaluation metric for this task is accuracy.

\paragraph{MSRPC} is a dataset of sentence pairs annotated for semantic
equivalence~\citep{dolan-brockett-2005-automatically} -- in GLUE, this
task is framed as a \emph{binary classification}.
The base prompt $p_k =$ ``\emph{mrpc sentence 1: \ldots \; sentence 2: \ldots}''
and $r_k=$ ``\emph{not\_equivalent}'' if the sentences are not equivalent,
and``\emph{equivalent}'' otherwise.
The evaluation metric for this task is accuracy.

\paragraph{STS-B} is a collection of sentence pairs annotated with a similarity
score from 1 to 5~\citep{cer-etal-2017-semeval}.
This task is a \emph{regression} task -- a model is tasked with predicting the similarity
score of the two input sentences.
The base prompt $p_k$ = ``\emph{STS sentence 1: \ldots  sentence 2 \ldots}''.
$r_k$ simply rounds the score of each input to 1 decimal point.
The evaluation metric for this task is the Spearman correlation coefficient.

\paragraph{QQP} is a dataset of question pairs from the website Quora, annotated
for whether or not the questions are semantically equivalent.
%\footnote{\url{https://data.quora.com/First-Quora-Dataset-Release-Question-Pairs}}
The task is a \emph{binary classification} task.
The base prompt $p_k$ = ``\emph{qqp question 1: \ldots question 2: \ldots}''
and $r_k$ is ``\emph{not duplicate} for non-equivalent questions, and
``\emph{duplicate}'' otherwise.
The evaluation metric for this task is accuracy.

\paragraph{MNLI \& MNLI-mm} is a dataset of sentence pairs with textual
entailment annotations~\citep{mnli-williams2018broadcoverage}.
The task is to determine whether the second sentence (the hypothesis) is
entailed, contradicted, or neither by the first sentence (the premise).
The task is a \emph{3-class classification} task.
The base prompt $p_k=$ ``\emph{MNLI premise: \ldots hypothesis: \ldots},
and $r_k$ is either ``\emph{entailment}'', ``\emph{neutral}'', or ``\emph{contradiction}''.
The evaluation metric for this task is accuracy.
The task comes with an additional mismatched test-set, MNLI-mm,
which contains inputs from different domains than the
training domain.

\paragraph{QNLI} is an entailment dataset constructed from
SQUAD~\citep{squad-rajpurkar-etal-2016-squad} by generating
question-sentence pairs from a question, and each sentence in
it's corresponding context.
The task is to determine whether to context sentence contains
the answer to the question, making the task a \emph{binary classification}
task.
Our base prompt $p_k=$ ``\emph{qnli question: \ldots sentence: \ldots}'',
and $r_k$ is ``\emph{entailment}'' if the context contains the answer,
and ``\emph{not entailment}'' otherwise.
The evaluation metric for this task is accuracy.

\paragraph{RTE} is composed of a series of textual entailment datasets
constructed from news and Wikipedia text. The task is to determine
whether sentence 2 is entailed by sentence 1, making the task a
\emph{binary classification} task.
The base prompt $p_k=$ ``\emph{rte sentence 1: \ldots sentence 2: \ldots}''
and $r_k$ is ``\emph{entailment}'' is sentence 2 is entailed and ``\emph{not entailment}''
otherwise.
The evaluation metric for this task is accuracy.

\subsection{DecaNLP}

The NLP Decathlon~\citep[DecaNLP;][]{McCann2018decaNLP}
frames 9 NLP tasks as question-answering problems, and proposes
a novel end-to-end Q\&A architecture (MQAN) to solve them.
We adopt 8 of the 9 tasks for our work, which we list and briefly
describe below.
Unlike GLUE, DecaNLP's framing of all tasks as questions results
in some tasks naturally being "self-specified" in the text-to-text
framework; for example, in SQUAD, the ``question'' portion of the
input specified to the model what task it should perform. Thus,
no prompt $p_k$ is necessary for SQUAD -- the question specifies
the task.

\paragraph{SST}
is the task of classifying whether or
not a given sentence has a positive
sentiment~\citep{sst-socher-etal-2013-recursive}, and also
appears in GLUE.
This task is a \emph{binary classification} task.
In DecaNLP the base prompt $p_k$ is \emph{`` \ldots Is this review negative or positive?''},
and $r_k$ is \emph{``positive''} for positive reviews and \emph{``negative''} otherwise.
The evaluation metric for this task is accuracy.

\paragraph{MNLI}
is a dataset of sentence pairs with textual
entailment annotations~\citep{mnli-williams2018broadcoverage}, and
also appears in GLUE.
The task is to determine whether the second sentence (the hypothesis) is
entailed, contradicted, or neither by the first sentence (the premise).
The task is a \emph{3-class classification} task.
The base prompt $p_k=$ ``\emph{Context: \ldots Premise: \ldots -- entailment, neutral, or contradiction?},
and $r_k$ is either ``\emph{entailment}'', ``\emph{neutral}'', or ``\emph{contradiction}''.
The evaluation metric for this task is accuracy.

\paragraph{IWSLT}
is a machine translation dataset - we leverage specifically the
English to German IWSLT 2016 task~\citep{iwslt-Cettolo2016TheI2}.
This task is a \emph{sequence to sequence} task.
Our base prompt is $p_k=$
\emph{``\ldots Translate this sentence into german''},
and $r_k$ is the identity map.
The evaluation metric for this task is BLEU.

\paragraph{CNN / Dailymail} is a summarization dataset of news
articles~\citep{cnn-nallapati-etal-2016-abstractive}.
The task is two generate an abstractive summary of a given input,
making this a \emph{sequence to sequence} task.
Our base prompt $p_k=$ \emph{``\ldots What is the summary?}'',
and $r_k$ is the identity map.
The evaluation metric for this task is ROUGE.

\paragraph{Seq2SQL} is a semantic parsing dataset,
where the task is to convert a natural language request
into a SQL query given some additional dataset
information~\citep{seq1seql-zhong2018seqsql}.
This is a \emph{sequence to sequence} task.
Our base prompt $p_k=$ \emph{``\ldots What is the translation 
from English to SQL?}'',
and $r_k$ is the identity map.
The evaluation metric for this task is Logical EM, where query outputs
are converted back into database logic and evaluate at that level.

\paragraph{SQUAD} is an extractive question answering
dataset~\citep{squad-rajpurkar-etal-2016-squad}.
For each example in squad, a context and question are
given -- the question is over the given context,
and the answer to the question exists as a span
inside the context.
Thus, this task is a \emph{span-labeling task}.
The task is specified by the question, so we use no
prompt for this task.
$r_k$ is simply a conversion of the span start and end points
into the natural language sentence that they cover.
The evaluation metric for this task is nF1.

\paragraph{QA-SRL}
is a semantic role labeling dataset -- 
semantic role labeling is the task of assigning semantic roles,
such as agent, goal, or result, to constituents of a sentence.
QA-SRL~\citep{qasrl-he-etal-2015-question} frames this task
as a question and answering problem, where the question specifies
a semantic role and asks which constituent fulfills that role,
and the answer is a span of the text.
Thus, this task is a \emph{span-labeling task}.
The task is specified by the question, so we use no
prompt for this task.
$r_k$ is simply a conversion of the span start and end points
into the natural language sentence that they cover.
The evaluation metric for this task is nF1.

\paragraph{QA-ZRE}
is a relation extraction dataset -- the task of extracting relationships between
one or more entities of a sentence.
Similar to QA-SRL, QA-ZRE~\citep{qazre-levy-etal-2017-zero} formulates
this problem as a question-answer dataset.
Here the question specifies a relation, and perhaps an entity, and
then asks what other entity in the sentence matches this relationship.
The answer can either be a span from the sentence, or that no entity
fulfills that relationship.
Thus, this task is a \emph{span-labeling task}.
The task is specified by the question, so we use no
prompt for this task.
$r_k$ is simply a conversion of the span start and end points
into the natural language sentence that they cover, or ``unanswerable''
if no entity fulfills the relationship.
The evaluation metric for this task is corpus-level F1.

\paragraph{Wino}
is a pronoun resolution dataset~\citep{winograd-10.5555/3031843.3031909},
with the task being to identify which entity is being referred to by a
specific pronoun in the sentence.
In DecaNLP this dataset is framed as a Q\&A problem task, where
the question simply asks which entity in the context is being
referred to by the ambiguous pronoun.
In the canonical setting this can be done by labeling the
entity in the text which is being referred to, yielding a
\emph{span-labeling task}.
The task is specified by the question, so we use no
prompt for this task.
$r_k$ is simply a conversion of the span start and end points
into the natural language sentence that they cover, or ``unanswerable''
if no entity fulfills the relationship.
The evaluation metric for this task is EM accuracy.

\section{Re-Initializing Sequence Decoders}
\label{app:no-pretraining-decoder}

\begin{figure}[t!]
    \centering
    \begin{subfigure}[t]{0.49\textwidth}
    \includegraphics[scale=0.45,trim=40px 0 0 30px]{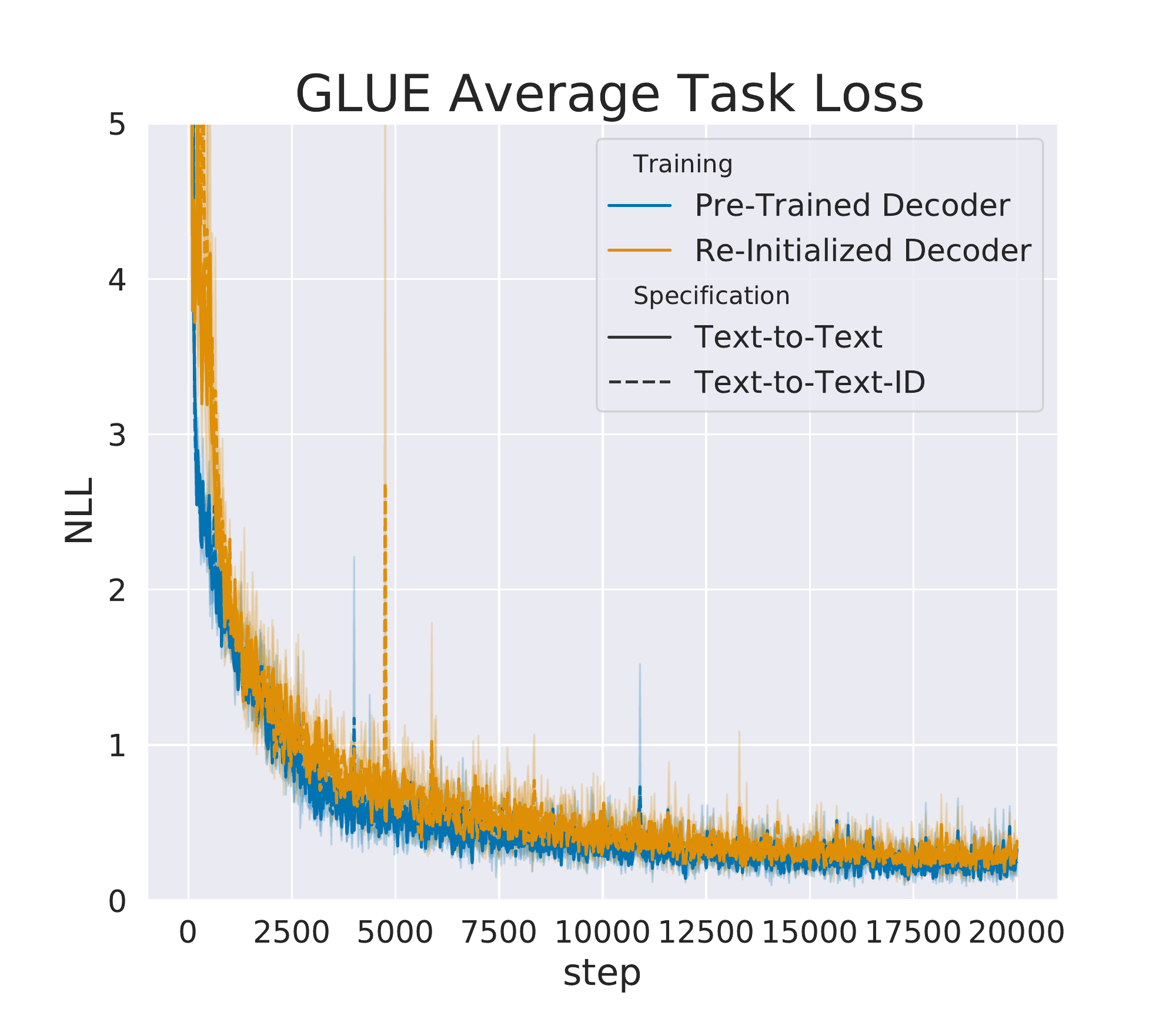}
    \end{subfigure}
    \hfill
    \begin{subfigure}[t]{.49\textwidth}
    \includegraphics[scale=0.45,trim=40px 0 0 30px]{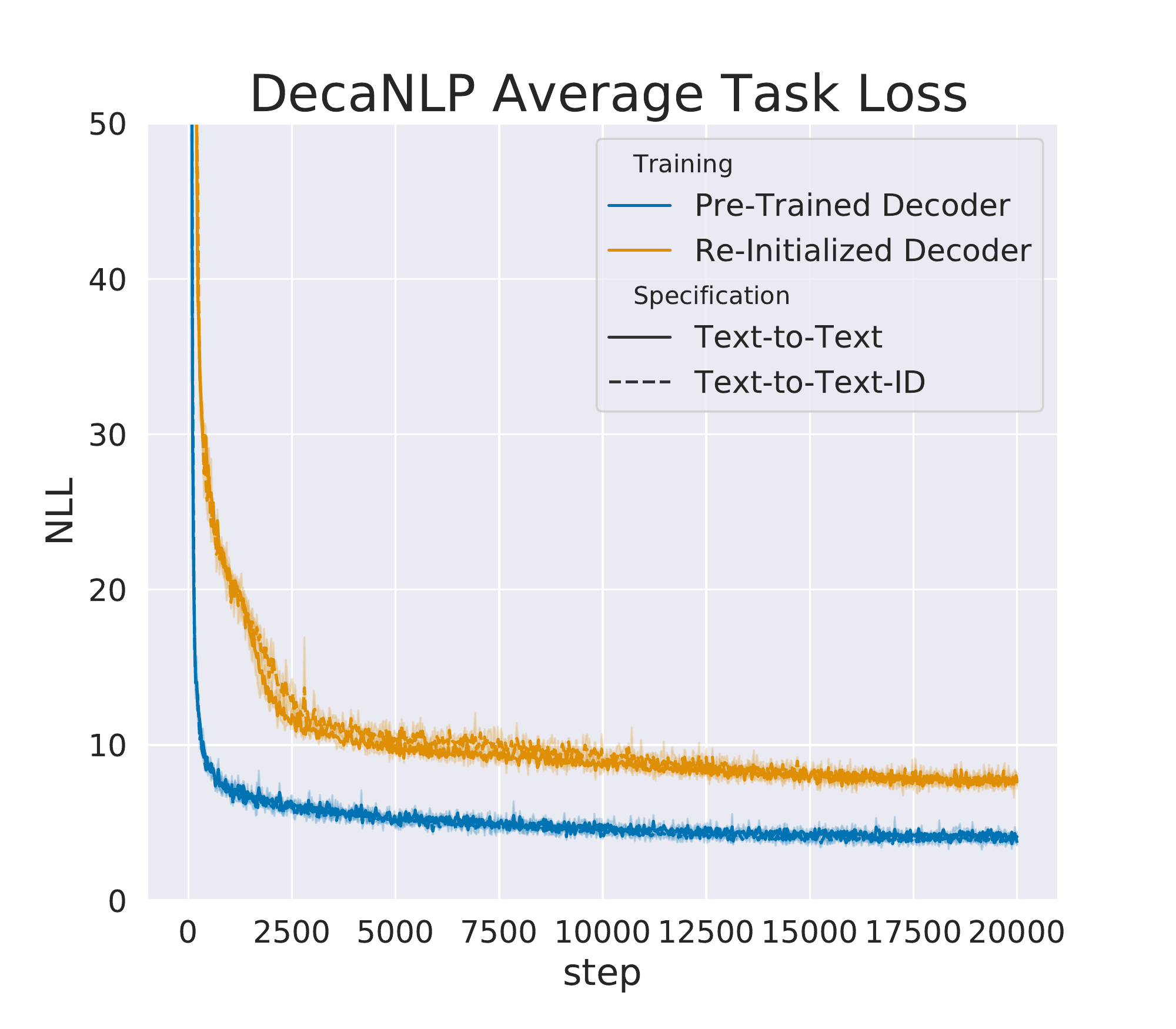}
    \end{subfigure}
    % \begin{subfigure}[t]{.49\textwidth}
    % \caption{DecaNLP}
    % \includegraphics[scale=0.28,trim=40px 80px 0 70px]{deca-neg-transfer.pdf}
    % \end{subfigure}
	\caption{\label{fig:reinit-loss} 
		The average task training loss of GLUE (top) and DecaNLP (bottom) for
		Text-to-Text and Text-to-Text-ID architectures, using pre-trained and
		re-initialized decoders.
		For GLUE, re-initializing the decoder has little effect on the
		convergence of text-to-text models: all models converge to equivalently
		optimal minima.
		However, for DecaNLP, re-initialized decoders are unable to fit the data
		as well as pre-trained decoders.
	}
\end{figure}

\begin{table*}[t!]
	\begin{center}
		\scalebox{0.71}{
\begin{tabular} {l l | c c c c c c c c c| c}

\toprule
\multicolumn{11}{c}{\textit{GLUE}} \\
Architecture & Decoder & CoLA & SST & MSRPC & STSB & QQP & MNLI & MNLI-mm & QNLI & RTE & {\bf Avg} \\
\midrule
Text-to-Text & Pre-Trained & 35.12 & 90.71 & 85.21 & 67.92 & 87.51 & 78.98 & 79.73 & 86.30 & 68.47 & 75.55 \\
Text-to-Text & Re-Initialized &  36.89 \cellcolor{green!17.66} & 90.37 \cellcolor{red!3.44} & 83.43 \cellcolor{red!17.81} & 67.85 \cellcolor{red!0.66} & 86.35 \cellcolor{red!11.57} & 77.62 \cellcolor{red!13.63} & 78.37 \cellcolor{red!13.66} & 85.07 \cellcolor{red!12.28} & 67.73 \cellcolor{red!7.46} & 74.85 \cellcolor{red!6.98} \\
\midrule
Text-to-Text-ID & Pre-Trained & 28.64 & 90.11 & 84.62 & 62.08 & 86.91 & 77.44 & 78.14 & 84.72 & 65.70 & 73.15 \\
Text-to-Text-ID & Re-Initialized &  38.32 \cellcolor{green!96.85} & 90.55 \cellcolor{green!4.42} & 84.31 \cellcolor{red!3.06} & 68.07 \cellcolor{green!59.89} & 86.54 \cellcolor{red!3.71} & 77.89 \cellcolor{green!4.50} & 78.57 \cellcolor{green!4.28} & 85.86 \cellcolor{green!11.38} & 68.74 \cellcolor{green!30.32} & 75.43 \cellcolor{green!22.76} \\
\midrule
\multicolumn{11}{c}{\textit{DecaNLP}} \\
Architecture & Decoder & SST & MNLI & IWSLT & CNN/DM & Seq2SQL & SQUAD & QA-SRL & QA-ZRE & Wino & {\bf Avg} \\
\midrule
Text-to-Text & Pre-Trained & 90.30 & 79.98 & 22.05 & 39.63 & 57.86 & 72.72 & 74.05 & 59.51 & 48.81 & 60.55 \\
Text-to-Text & Re-Initialized &  88.98 \cellcolor{red!4.63} & 75.51 \cellcolor{red!15.65} & 9.22 \cellcolor{red!44.89} & 27.51 \cellcolor{red!42.43} & 42.01 \cellcolor{red!55.45} & 66.88 \cellcolor{red!20.45} & 56.11 \cellcolor{red!62.79} & 53.45 \cellcolor{red!21.22} & 22.02 \cellcolor{red!93.75} & 49.08 \cellcolor{red!40.14} \\

\midrule
Text-to-Text-ID & Pre-Trained & 91.61 & 81.10 & 23.05 & 39.88 & 58.52 & 72.53 & 62.90 & 55.57 & 31.85 & 57.44 \\
Text-to-Text-ID & Re-Initialized &  90.04 \cellcolor{red!5.49} & 75.32 \cellcolor{red!20.21} & 6.40 \cellcolor{red!58.27} & 28.37 \cellcolor{red!40.27} & 35.22 \cellcolor{red!81.57} & 61.87 \cellcolor{red!37.31} & 2.76 \cellcolor{red!99.99} & 36.92 \cellcolor{red!65.27} & 0.79 \cellcolor{red!99.99} & 37.52 \cellcolor{red!69.73} \\

\bottomrule
\end{tabular}
}
\end{center}
\caption{\label{table:non-pretrained}
	The effects of re-initializing the decoder head versus leveraging it's pre-trained parameters.
	We see that, for GLUE, re-initializing the decoder head has little effect on performance;
	because of the high overlap resulting from a fixed $r_k$ for classification tasks, a strong
	language model is not necessary to achieve competitive results on GLUE.
	However, for tasks whose output spaces require generalization to {\it natural language}
	(most tasks for DecaNLP), a strong language model is necessary.
	Re-initializing the decoder heads for DecaNLP results in a catastrophic decrease in performance.
}
\end{table*}

When studying the effects of Factor \fact{1} in \autoref{sec:fact-1-2}, we
leverage pre-trained T5 decoders as the replacement for each task-specific
head.
Using T5 decoders as task-specific heads moves all task output spaces into
$\mathcal{X}$, and ensures that all tasks can be trained with the same loss
function ($\mathcal{L}_{LM}$).
However, pre-training is {\it not} necessary for this shift, and could be
considered a confounding factor in moving from canonical to text-to-text models,
as most canonical heads do not benefit from pre-training.
Here, we explore the effects of {\it re-initializing} the T5 decoder heads,
removing the biases of pre-training when modeling the output space of each
task.
In \autoref{table:non-pretrained} we plot multi-task model performance on
both GLUE and DecaNLP using Pre-trained and Re-Initialized decoder heads, for
both Text-to-Text and Text-to-Text-ID architectures.
We additionally plot the average task loss during training of all models in
\autoref{fig:reinit-loss}.

Perhaps unsurprisingly, the effects of re-initializing the decoder on GLUE are
small: in both Text-to-Text and Text-to-Text-ID models, performance is
similar when using re-initialized or pre-trained decoders.
Two notable exceptions are CoLA and STS-B, both regression-like tasks;
when trained with independent heads, performance is {\it significantly} higher
when using a re-initialized decoder.
This result may suggest that the biases of a pre-trained decoder head are
ill-suited for regression-like tasks.
Interestingly, when task-heads are combined across tasks this discrepancy largely
disappears: Text-to-Text architectures outperform Text-to-Text-ID architectures
when using pre-trained decoders.
Overall, the effects of re-initialization on performance are reflected in the
training trajectories: in \autoref{fig:reinit-loss} (top) we see that
re-initializing the decoder has almost no effect on model convergence for GLUE.

However, the results of re-initializing the decoder on DecaNLP are significantly
negative.
7 of the 9 tasks in DecaNLP are span-labeling or sequence generation tasks;
unlike classification tasks, these 7 tasks contain extremely diverse output
spaces, requiring either extractive or abstractive sequence generation.
Because good generalization on these tasks requires a strong language model,
re-initializing the decoder on this benchmark is disastrous for most tasks
considered; not only is performance significantly affected, but we can see from
\autoref{fig:reinit-loss} that re-initialized decoders fail to even converge to
good minima.
The only exceptions are the SST and MNLI tasks, which are both classification
tasks.

\section{Increasing Model Capacity}
\label{app:larger-model}

\begin{table*}[t!]
	\begin{center}
		\scalebox{0.75}{
\begin{tabular} {l l | c c c c c c c c c| c}
\toprule
\multicolumn{11}{c}{\textit{GLUE}} \\
%\midrule
Architecture & Decoder & CoLA & SST & MSRPC & STSB & QQP & MNLI & MNLI-mm & QNLI & RTE & {\bf Avg} \\
\midrule
Canonical & Single-Task & 42.93 & 92.49 & 81.91 & 88.69 & 90.04 & 84.97 & 85.56 & 89.68 & 64.33 & 78.76 \\
Canonical & Multi-Task &  40.52 \cellcolor{red!24.03} & 92.16 \cellcolor{red!3.25} & 84.89 \cellcolor{green!29.74} & 88.09 \cellcolor{red!6.04} & 89.27 \cellcolor{red!7.71} & 82.49 \cellcolor{red!24.86} & 83.15 \cellcolor{red!24.10} & 88.93 \cellcolor{red!7.48} & 70.28 \cellcolor{green!59.45} & 79.97 \cellcolor{green!12.16} \\
\midrule
Text-to-Text & Single-Task & 43.75 & 92.16 & 82.21 & 66.48 & 90.11 & 85.45 & 85.76 & 90.86 & 63.90 & 77.56 \\
Text-to-Text & Multi-Task &  45.83 \cellcolor{green!20.77} & 92.59 \cellcolor{green!4.36} & 86.91 \cellcolor{green!47.06} & 67.92 \cellcolor{green!14.35} & 89.45 \cellcolor{red!6.60} & 84.12 \cellcolor{red!13.33} & 84.76 \cellcolor{red!10.01} & 89.85 \cellcolor{red!10.10} & 73.36 \cellcolor{green!94.58} & 79.42 \cellcolor{green!18.60} \\
\bottomrule
\end{tabular}
}
\end{center}
\caption{\label{table:t5-base}
	Negative transfer between text-to-text and canonical models with the T5-Base
	architecture (a larger model than used in the main paper), averaged over
	5 random seeds.
	T5-Base largely corroborates our findings, exhibiting similar amounts of
	negative \& positive transfer across canonical and text-to-text architectures.
}
\end{table*}

Finally, we explore the effects of increasing model size on our GLUE results.
All experiments in the main paper use the T5-Small architecture, which contains
far fewer parameters than current SotA text-to-text models.
As a small corroborative experiment, we study how expanding the model size from
T5-Small to T5-Base affects our main results.

In \autoref{table:t5-base} we replicate our core experiment (studying negative
transfer across Canonical \& Text-to-Text architectures) using a pre-trained
T5-Base model as the base of our architectures.
We find that increasing the model size does not have a significant effect on our
key conclusion: negative transfer at the task-level remains similar in
canonical and text-to-text settings.
However, as noted by \citet{firat-mtl}, we find that increasing the model size
alleviates negative transfer on the whole: T5-Small exhibits negative transfer
at the benchmark level for GLUE (\autoref{table:main}), whereas T5-Base 
benefits from positive transfer when considering the average task performance.

\end{document}